\documentclass[acmlarge,screen]{acmart}

\def\BibTeX{{\rm B\kern-.05em{\sc i\kern-.025em b}\kern-.08emT\kern-.1667em\lower.7ex\hbox{E}\kern-.125emX}}

\copyrightyear{2020}
\acmYear{2020}
\setcopyright{acmlicensed}

\begin{document}

\title{Deep Learning Based Text Classification: A Comprehensive Review}

%

\author{S\lowercase{hervin} M\lowercase{inaee}}
\email{sminaee@snapchat.com,}
\affiliation{\textit{Snapchat Inc}}

\author{N\lowercase{al} K\lowercase{alchbrenner}}
\email{nalk@google.com,}
\affiliation{\textit{Google Brain, Amsterdam}}

\author{E\lowercase{rik} C\lowercase{ambria}}
\email{cambria@ntu.edu.sg,}
\affiliation{\textit{Nanyang Technological University, Singapore}}

\author{N\lowercase{arjes} N\lowercase{ikzad}}
\email{narjes.nikzad@tabrizu.ac.ir,}
\affiliation{\textit{University of Tabriz}}

\author{M\lowercase{eysam} C\lowercase{henaghlu}}
\email{m.asgari.c@tabrizu.ac.ir,}
\affiliation{\textit{University of Tabriz}}

\author{J\lowercase{ianfeng} G\lowercase{ao}}
\email{jfgao@microsoft.com,}
\affiliation{\textit{Microsoft Research, Redmond}}

%

%
\begin{abstract}
\textbf{Abstract.}
Deep learning based models have surpassed classical machine learning based approaches in various text classification tasks, including sentiment analysis, news categorization, question answering, and natural language inference.
In this paper, we provide a comprehensive review of more than 150 deep learning based models for text classification developed in recent years, and discuss their technical contributions, similarities, and strengths.
We also provide a summary of more than 40 popular datasets widely used for text classification. 
Finally, we provide a quantitative analysis of the performance of different deep learning models on popular benchmarks, and discuss future research directions.
\end{abstract}

%
%

%


%
\maketitle

\section{\textbf{Introduction}}

Text classification, also known as text categorization, is a classical problem in natural language processing (NLP), which aims to assign labels or tags to textual units such as sentences, queries, paragraphs, and documents. 
It has a wide range of applications including question answering, spam detection, sentiment analysis, news categorization, user intent classification, content moderation, and so on.
Text data can come from different sources, including web data, emails, chats, social media, tickets, insurance claims, user reviews, and questions and answers from customer services, to name a few.
Text is an extremely rich source of information. 
But extracting insights from text can be challenging and time-consuming, due to its unstructured nature.

Text classification can be performed either through manual annotation or by automatic labeling. 
With the growing scale of text data in industrial applications, automatic text classification is becoming increasingly important.
Approaches to automatic text classification can be grouped into two categories:
\begin{itemize}
  \item Rule-based methods
  \item Machine learning (data-driven) based methods
\end{itemize}
Rule-based methods classify text into different categories using a set of pre-defined rules, and  require a deep domain knowledge.
On the other hand, machine learning based approaches learn to classify text based on observations of data. 
Using pre-labeled examples as training data, a machine learning algorithm learns inherent associations between texts and their labels. 

Machine learning models have drawn lots of attention in recent years. Most classical machine learning based models follow the two-step procedure. In the first step, some hand-crafted features are extracted from the documents (or any other textual unit). In the second step, those features are fed to a classifier to make a prediction.
Popular hand-crafted features include bag of words (BoW) and their extensions.
Popular choices of classification algorithms include Na\"ive Bayes, support vector machines (SVM), hidden Markov model (HMM), gradient boosting trees, and random forests.
The two-step approach has several limitations. For example, reliance on the hand-crafted features 
requires tedious feature engineering and analysis to obtain good performance. 
In addition, the strong dependence on domain knowledge for designing 
features makes the method difficult to generalize to new tasks. 
Finally, these models cannot take full advantage of large amounts of training data because the features (or feature templates) are pre-defined.


Neural approaches have been explored to address the limitations due to the use of hand-craft features. The core component of these approaches is a machine-learned embedding model that maps text into a low-dimensional continuous feature vector, thus no hand-crafted features is needed. 
One of earliest embedding models is latent semantic analysis (LSA) developed by Dumais et al. \cite{deerwester1990indexing} in 1989. LSA is a linear model with less than 1 million parameters, trained on 200K words. 
In 2001, Bengio et al. \cite{bengio2003neural} propose the first neural language model based on a feed-forward neural network trained on 14 million words. However, these early embedding models underperform classical models using hand-crafted features, and thus are not widely adopted.
A paradigm shift starts when much larger embedding models are developed using much larger amounts of training data. 
In 2013, Google develops a series of word2vec models \cite{mikolov2013distributed} that are trained on 6 billion words and immediately become popular for many NLP tasks.
In 2017, the teams from AI2 and University of Washington develops a contextual embedding model based on a 3-layer bidirectional LSTM with 93M parameters trained on 1B words. The model, called ELMo \cite{peters2018deep}, works much better than word2vec because they capture contextual information.
In 2018, OpenAI starts building embedding models using Transformer \cite{vaswani2017attention}, a new NN architecture developed by Google. Transformer is solely based on attention which substantially improves the efficiency of large-scale model training on TPU. Their first model is called GPT \cite{radford2018improving}, which is now widely used for text generation tasks. The same year, Google develops BERT \cite{devlin2018bert} based on bidirectional transformer. BERT consists of 340M parameters, trained on 3.3 billion words, and is the current state of the art embedding model.
The trend of using larger models and more training data continues. By the time this paper is published, OpenAI's latest GPT-3 model \cite{brown2020language} contains 170 billion parameters, and Google's GShard \cite{lepikhin2020gshard} contains 600 billion parameters. 

Although these gigantic models show very impressive performance on various NLP tasks, some researchers argue that they do not really understand language and are not robust enough for many mission-critical domains \cite{marcus2019rebooting,marcus2020next,nie2019adversarial,jin2019bert,liu2020adversarial}. Recently, there is an growing interest in exploring neuro-symbolic hybrid models (e.g., \cite{andreas2016learning,iyyer2017search,schlag2019enhancing,gao2020robust}) to address some of the fundamental limitations of neural models, such as lack of grounding, being unable to perform symbolic reasoning, not interpretable. These works, although important, are beyond the scope of this paper.

While there are many good reviews and text books on text classification methods and applications in general e.g., ~\cite{kowsari2019text,manning2008introduction,jurasky2008speech}, this survey is unique in that it presents a comprehensive review on more than 150 deep learning (DL) models 
developed for various text classification tasks, including sentiment analysis, news categorization, topic classification, question answering (QA), and natural language inference (NLI), over the course of the past six years.
In particular, we group these works into several categories based on their neural network architectures, including recurrent neural networks (RNNs), convolutional neural networks (CNNs),  attention, Transformers, Capsule Nets, and so on.
The contributions of this paper can be summarized as follows:
\begin{itemize}
  \item We present a detailed overview of more than 150 DL models proposed for text classification.
  \item We review more than 40 popular text classification datasets.
  \item We provide a quantitative analysis of the performance of a selected set of DL models on 16 popular benchmarks.
  \item We discuss remaining challenges and future directions. 
\end{itemize}

\subsection{\textbf{Text Classification Tasks}}

Text Classification (TC) is the process of categorizing texts (e.g., tweets, news articles, customer reviews) into organized groups. 
Typical TC tasks include sentiment analysis, news categorization and topic classification.
Recently, researchers show that it is effective to cast many natural language understanding (NLU) tasks (e.g., extractive question answering, natural language inference) as TC by allowing DL-based text classifiers to take a pair of texts as input (e.g., \cite{wang2018glue,devlin2018bert,liu2019multi}).  
This section introduces five TC tasks discussed in this paper, including three typical TC tasks and two NLU tasks that are commonly cast as TC in many recent DL studies.  

\paragraph{\textbf{Sentiment Analysis}}
This is the task of analyzing people's opinions in textual data (e.g., product reviews, movie reviews, or tweets), and extracting their polarity and viewpoint.
The task can be cast as either a binary or a multi-class problem. 
Binary sentiment analysis classifies texts into positive and negative classes, while multi-class sentiment analysis classifies texts into fine-grained labels or multi-level intensities.

\paragraph{\textbf{News Categorization}}

News contents are among the most important information sources.
A news classification system helps users obtain information of interest in real-time by 
e.g., identifying emerging news topics or recommending relevant news based on user interests.

\paragraph{\textbf{Topic Analysis}}
The task, also known as \emph{topic classification}, aims to identify the theme or topics of a text (e.g., whether a product review is about ``customer support'' or ``ease of use'').  

\paragraph{\textbf{Question Answering (QA)}}

There are two types of QA tasks: extractive and generative.
Extractive QA is a TC task: Given a question and a set of candidate answers (e.g., text spans in a document in SQuAD~\cite{rajpurkar2016squad}), a system classifies each candidate answer as correct or not. 
Generative QA is a text generation task since it requires generating answers on the fly.
This paper only discusses extractive QA.

\paragraph{\textbf{Natural language inference (NLI)}}
NLI, also known as \textit{recognizing textual entailment} (RTE), predicts whether the meaning of one text can be inferred from another. 
An NLI system needs to assign to a pair of text units a label such as entailment, contradiction, and neutral~\cite{marelli2014semeval}. 
Paraphrasing is a generalized form of NLI, also known as \textit{text pair comparison}, the task of measuring the semantic similarity of a sentence pair indicating how likely one sentence is a paraphrase of the other. 

\subsection{\textbf{Paper Structure}}
The rest of the paper is structured as follows:
Section~\ref{sec:Deep_text} presents a comprehensive review of more than 150 DL-based text classification models. 
Section~\ref{sec:choose} presents a recipe of building text classifiers using DL models.
Section~\ref{sec:datasets}
reviews some of the most popular TC datasets. 
Section~\ref{sec:performance} presents a quantitative performance analysis of a selected set of DL models on 16 benchmarks.
Section~\ref{sec:challenges} discusses the main challenges and future directions for DL-based TC methods.
Section~\ref{sec:conclusions} concludes the paper.



\section{\textbf{Deep Learning Models for Text Classification}}
\label{sec:Deep_text}
This section reviews more than 150 DL models proposed for various TC tasks. 
For clarify, we group these models into several categories based on their model architectures\footnote{These categories are introduced mainly for a pedagogical purpose. They are by no means exclusive to each other. For example, the Transformer uses a composite structure consisting of feed-forward layers and the attention mechanism, and memory-augment networks also involve the attention mechanism.}:
\begin{itemize}
  \item Feed-forward networks view text as a bag of words (Section~\ref{subsec:feed-forward-neural-networks}).
  \item RNN-based models view text as a sequence of words, and are intended to capture word dependencies and text structures (Section~\ref{subsec:rnn-based-models}).
  \item CNN-based models are trained to recognize patterns in text, such as key phrases, for TC (Section~\ref{subsec:cnn-based-models}).
  \item Capsule networks address the information loss problem suffered by the pooling operations of CNNs, and recently have been applied to TC (Section~\ref{subsec:capsule}).
  \item The attention mechanism is effective to identify correlated words in text, and has become a useful tool in developing DL models (Section~\ref{subsec:attention}).
  \item Memory-augmented networks combine neural networks with a form of external memory, which the models can read from and write to (Section~\ref{subsec:memory-networks}). 
  \item Graph neural networks are designed to capture internal graph structures of natural language, such as syntactic and semantic parse trees (Section~\ref{subsec:gnn}). 
  \item Siamese Neural Networks are designed for text matching, a special case of TC (Section~\ref{subsec:siamese-networks}) . 
  \item Hybrid models combine attention, RNNs, CNNs, etc. to capture local and global features of sentences and documents (Section~\ref{subsec:hybrid-models}). 
  \item Transformers allow for much more parallelization than RNNs, making it possible to efficiently (pre-)train very big language models using GPUs (Section~\ref{subsec:transformers}). 
  \item Finally, in Section~\ref{subsec:beyond-supervised-learning}, we review modeling technologies that are beyond supervised learning, including unsupervised learning using autoencoder and adversarial training, and reinforcement learning.
\end{itemize}

Readers are expected to be reasonably familiar with basic DL models to comprehend the content of this section.
Readers are referred to the DL textbook by Goodfellow et al.~\cite{goodfellow2016deep} for more details.

\subsection{\textbf{Feed-Forward Neural Networks}}
\label{subsec:feed-forward-neural-networks}

\begin{figure}[h]
\begin{center}
  \includegraphics[page=1,width=0.6\linewidth]{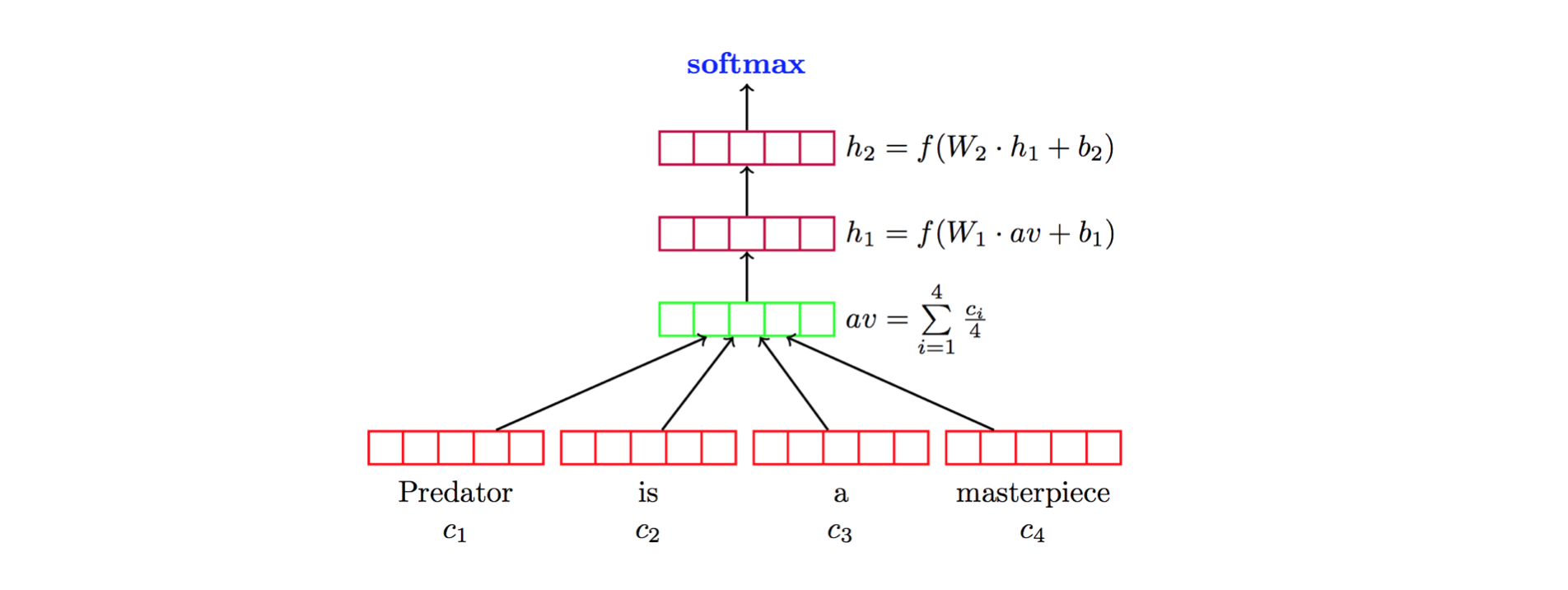}
\end{center}
  \caption{The architecture of the Deep Average Network (DAN)~\cite{iyyer2015deep}.}
\label{fig:DAN}
\end{figure}

Feed-forward networks are among the simplest DL models for text representation. Yet, they have achieved high accuracy on many TC benchmarks. 
These models view text as a bag of words. 
For each word, they learn a vector representation using an embedding model such as word2vec~\cite{mikolov2013efficient} or Glove~\cite{pennington2014glove}, take the vector sum or average of the embeddings as the representation of the text, pass it through one or more feed-forward layers, known as Multi-Layer Perceptrons (MLPs), and then perform classification on the final layer's representation using a classifier such as logistic regression, Na\"ive Bayes, or SVM~\cite{iyyer2015deep}. 
An example of these models is the Deep Average Network (DAN)~\cite{iyyer2015deep}, whose architecture is shown in Fig.~\ref{fig:DAN}. Despite its simplicity, DAN outperforms other more sophisticated models which are designed to explicitly learn the compositionality of texts. For example, DAN outperforms syntactic models on datasets with high syntactic variance. 
Joulin et al.~\cite{joulin2016fasttext} propose a simple and efficient text classifier called fastText. Like DAN, fastText views text as a bag of words. Unlike DAN, fastText uses a bag of n-grams as additional features to capture local word order information. This turns out to be very efficient in practice,  achieving comparable results to the methods that explicitly use the word order~\cite{wang2012baselines}. 

\begin{figure}[h]
\begin{center}
  \includegraphics[page=1,width=0.9\linewidth]{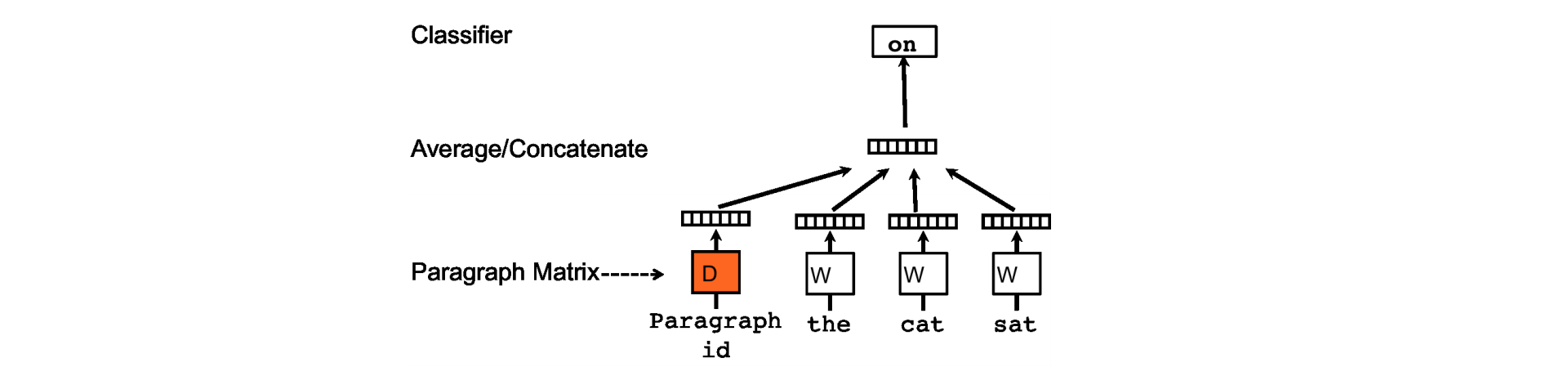}
\end{center}
  \caption{The doc2vec model~\cite{le2014distributed}.}
\label{fig:doc2vec}
\end{figure}

Le and Mikolov~\cite{le2014distributed} propose doc2vec, 
which uses an unsupervised algorithm to learn fixed-length feature representations of variable-length pieces of texts, such as sentences, paragraphs, and documents. 
As shown in Fig.~\ref{fig:doc2vec}, the architecture of doc2vec is similar to that of the Continuous
Bag of Words (CBOW) model \cite{mikolov2013efficient, mikolov2013distributed}. 
The only difference is the additional paragraph token that is mapped to a paragraph vector via matrix $D$. In doc2vec, the concatenation or average of this vector with a context of three words is used to predict the fourth word. The paragraph vector represents the missing information from the current context and can act as a memory of the topic of the paragraph. After being trained, the paragraph vector is used as features for the paragraph (e.g., in lieu of or in addition to BoW), 
and fed to a classifier for prediction. Doc2vec achieves new state of the art results on several TC tasks when it is published.

\subsection{\textbf{RNN-Based Models}}
\label{subsec:rnn-based-models}

RNN-based models view text as a sequence of words, and are intended to capture word dependencies and text structures for TC. However, vanilla RNN models do not perform well, and often underperform feed-forward neural networks. 
Among many variants to RNNs, Long Short-Term Memory (LSTM) is the most popular architecture, which is designed to better capture long term dependencies. LSTM addresses the gradient vanishing or exploding problems suffered by the vanilla RNNs by introducing a memory cell to remember values over arbitrary time intervals, and three gates (input gate, output gate, forget gate) to regulate the flow of information into and out of the cell.
There have been works on improving RNNs and LSTM models for TC by capturing richer information, such as tree structures of natural language, long-span word relations in text, document topics, and so on. 

\begin{figure}[h]
\begin{center}
  \includegraphics[page=1,width=0.8\linewidth]{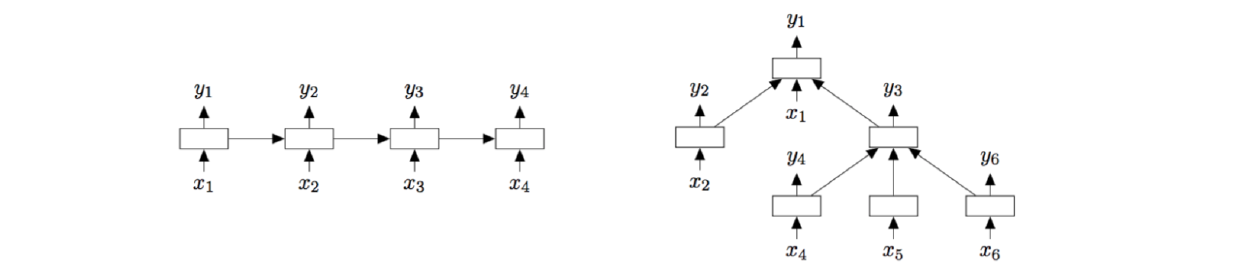}
\end{center}
  \caption{(Left) A chain-structured LSTM network and (right) a tree-structured LSTM network with arbitrary branching factor~\cite{tai2015improved}. Here $x_i$ and $y_i$ denote the input and output of each cell.}
\label{fig:tree_lstm}
\end{figure}

Tai et al.~\cite{tai2015improved} develop a Tree-LSTM model, a generalization of LSTM to tree-structured network typologies, to learn rich semantic representations. The authors argue that Tree-LSTM is a better model than the chain-structured LSTM for NLP tasks because natural language exhibits syntactic properties that would naturally combine words to phrases. They validate the effectiveness of Tree-LSTM on two tasks: sentiment classification and predicting the semantic relatedness of two sentences. The architectures of these models are shown in Fig.~\ref{fig:tree_lstm}. 
Zhu et al.~\cite{zhu2015long} also extend the chain-structured LSTM to tree structures, using a memory cell to store the history of multiple child cells or multiple descendant cells in a recursive process. They argue that the new model provides a principled way of considering long-distance interaction over hierarchies, e.g., language or image parse structures.

To model long-span word relations for machine reading, Cheng et al.~\cite{cheng2016long} augment the LSTM architecture with a memory network in place of a single memory cell. This enables adaptive memory usage during recurrence with neural attention, offering a way to weakly induce relations among tokens. This model achieves promising results on language modeling, sentiment analysis, and NLI.

The Multi-Timescale LSTM (MT-LSTM) neural network~\cite{liu2015multi}
is also designed to model long texts, such as sentences and documents, by capturing valuable information with different timescales. MT-LSTM partitions the hidden states of a standard LSTM model into several groups. Each group is activated and updated at different time periods. Thus, MT-LSTM can model very long documents. MT-LSTM is reported to outperform a set of baselines, including the models based on LSTM and RNN, on TC. 

RNNs are good at capturing the local structure of a word sequence, but face difficulties remembering long-range dependencies. In contrast, latent topic models are able to capture the global semantic structure of a document but do not account for word ordering. Dieng et al.~\cite{dieng2016topicrnn} propose a TopicRNN model to integrate the merits of RNNs and latent topic models. It captures local (syntactic) dependencies using RNNs and global (semantic) dependencies using latent topics. TopicRNN is reported to outperform RNN baselines for sentiment analysis. 


There are other interesting RNN-based models.
Liu et al.~\cite{liu2016recurrent} use multi-task learning to train RNNs to leverage labeled training data from multiple related tasks.
Johnson and Rie~\cite{johnson2016supervised} explore a text region embedding method using LSTM.
Zhou et al.~\cite{zhou2016text} integrate a Bidirectional-LSTM (Bi-LSTM) model with two-dimensional max-pooling to capture text features.
Wang et al.~\cite{wang2017bilateral} propose a bilateral multi-perspective matching model under the ``matching-aggregation'' framework.
Wan et al.~\cite{wan2016deep} explore semantic matching using multiple positional sentence representations generated by a bi-directional LSMT model.

It is worth noting that RNNs belong to a broad category of DNNs, known as \emph{recursive neural networks}. A recursive neural network applies the same set of weights recursively over a structure input to produce a structured prediction or a vector representation over variable-size input.  Whereas RNNs are recursive neural networks with a linear chain structure input, there are recursive neural networks that operate on hierarchical structures, such as parse trees of natural language sentences \cite{socher2013recursive}, combining child representations into parent representations. RNNs are the most popular recursive neural networks for TC because they are effective and easy to use -- they view text as a sequence of tokens without requiring additional structure labels such as parse trees.

\subsection{\textbf{CNN-Based Models}}
\label{subsec:cnn-based-models}

RNNs are trained to recognize patterns across time, whereas CNNs learn to recognize patterns across space~\cite{lecun1998gradient}. RNNs work well for the NLP tasks such as POS tagging or QA where the comprehension of long-range semantics is required, while CNNs work well where detecting local and position-invariant patterns is important. These patterns could be key phrases that express a particular sentiment like ``I like'' or a topic like ''endangered species''. Thus, CNNs have become one of the most popular model architectures for TC.

\begin{figure}[h]
\begin{center}
  \includegraphics[page=1,width=0.43\linewidth]{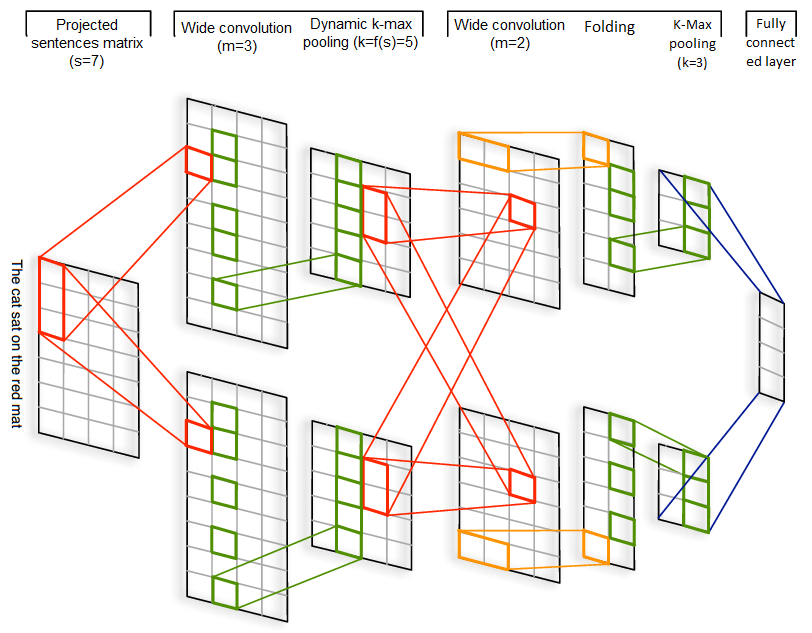}
\end{center}
  \caption{The architecture of DCNN model~\cite{Kalchbrenner2014}.}
\label{fig:dcnn}
\end{figure}

One of the first CNN-based models for TC is proposed by Kalchbrenner et al.~\cite{Kalchbrenner2014}. 
The model uses dynamic $k$-max-pooling, and is called the Dynamic CNN (DCNN). 
As illustrated in Fig.~\ref{fig:dcnn}, the first layer of DCNN constructs a sentence matrix using the embedding for each word in the sentence. Then a convolutional architecture that alternates wide convolutional layers with dynamic pooling layers given by dynamic $k$-max-pooling is used to generate a feature map over the sentence that is capable of explicitly capturing short and long-range relations of words and phrases. The pooling parameter $k$ can be dynamically chosen depending on the sentence size and the level in the convolution hierarchy.

\begin{figure}[h]
\begin{center}
  \includegraphics[page=1,width=0.6\linewidth]{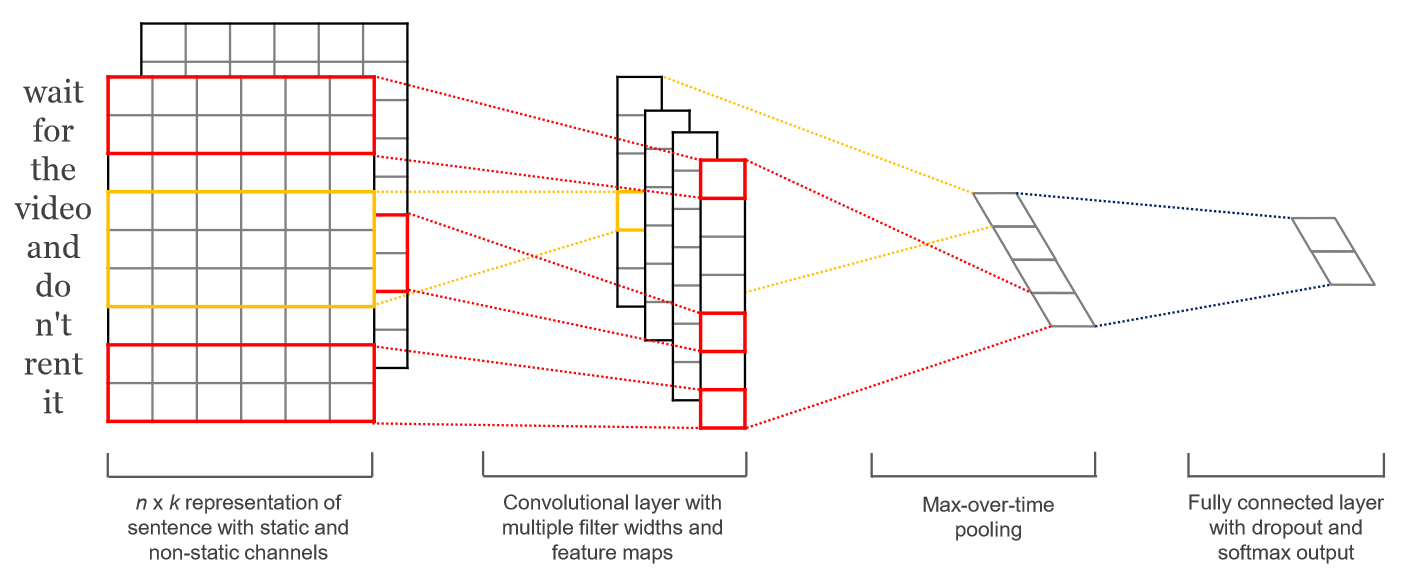}
\end{center}
  \caption{The architecture of a sample CNN model for text classification. courtesy of Yoon Kim~\cite{Kim2014}.}
\label{fig:kim}
\end{figure}

Later, Kim~\cite{Kim2014} proposes a much simpler CNN-based model than DCNN for TC. 
As shown in Fig.~\ref{fig:kim}, Kim's model uses only one layer of convolution on top of the word vectors obtained from an unsupervised neural language model i.e., word2vec. 
Kim also compares four different approaches to learning word embeddings: 
(1) CNN-rand, where all word embeddings are randomly initialized and then modified during training;
(2) CNN-static, where the pre-trained word2vec embeddings are used and stay fixed during model training;
(3) CNN-non-static, where the word2vec embeddings are fine-tuned during training for each task; and 
(4) CNN-multi-channel, where two sets of word embedding vectors are used, both are initialized using word2vec, with one updated during model training while the other fixed.
These CNN-based models are reported to improve upon the state of the art on sentiment analysis and question classification.

There have been efforts of improving the architectures of CNN-based models of~\cite{Kalchbrenner2014,Kim2014}.
Liu et al.~\cite{Liu2017} propose a new CNN-based model that makes two modifications to the architecture of Kim-CNN~\cite{Kim2014}. First, a dynamic max-pooling scheme is adopted to capture more fine-grained features from different regions of the document. Second, a hidden bottleneck layer is inserted between the pooling and output layers to learn compact document representations to reduce model size and boost model performance.
In~\cite{Johnson2015,Johnson2017}, instead of using pre-trained low-dimensional word vectors as input to CNNs, the authors directly apply CNNs to high-dimensional text data to learn the embeddings of small text regions for classification.

\begin{figure}[h]
\begin{center}
  \includegraphics[page=1,width=0.55\linewidth]{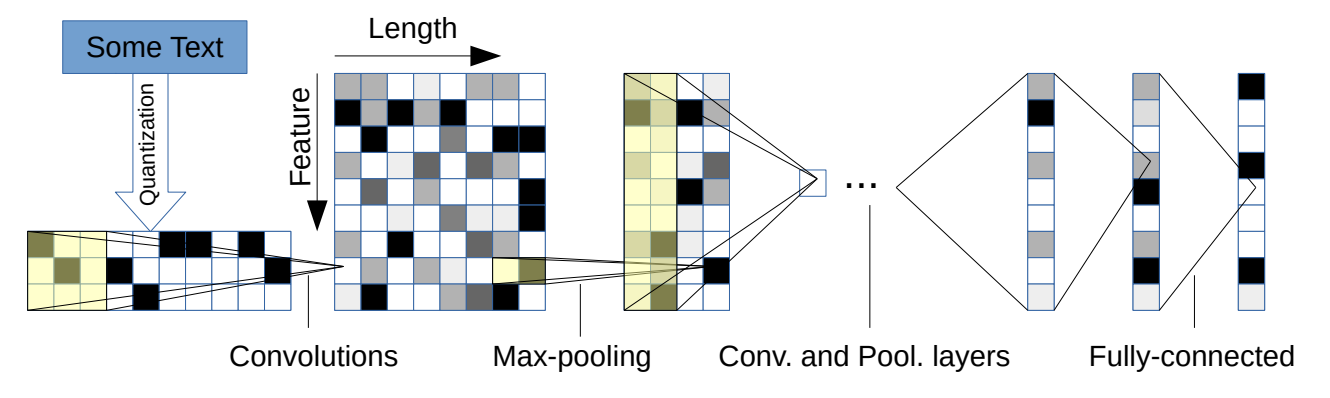}
\end{center}
  \caption{The architecture of a character-level CNN model~\cite{zhang2015character}.}
\label{CharaCNN}
\end{figure}

Character-level CNNs have also been explored for TC \cite{zhang2015character, kim2016character}. 
One of the first such models is proposed by Zhang et al.~\cite{zhang2015character}. 
As illustrated in Fig.~\ref{CharaCNN}, 
the model takes as input the characters in a fixed-sized, encoded as one-hot vectors, passes them through a deep CNN model that consists of six convolutional layers with pooling operations and three fully connected layers. 
Prusa et al.~\cite{Prusa2016} present an approach to encoding text using CNNs that greatly reduces memory consumption and training time required to learn character-level text representations. 
This approach scales well with alphabet size, allowing to preserve more information from the original text to enhance classification performance.

There are studies on investigating the impact of word embeddings and CNN architectures on model performance. 
Inspired by VGG~\cite{Simonyan2015} and ResNets~\cite{He2016}, Conneau et al.~\cite{conneau2016very} present a Very Deep CNN (VDCNN) model for text processing. It operates directly at the character level and uses only small convolutions and pooling operations. This study shows that the performance of VDCNN improves with the increase of the depth. 
Duque et al.~\cite{Duque2019} modify the structure of VDCNN to fit mobile platforms' constraints without much performance degradation. They are able to compress the model size by 10x to 20x with an accuracy loss between 0.4\% to 1.3\%.
Le et al.~\cite{le2018convolutional} show that deep models indeed outperform shallow models when the text input is represented as a sequence of characters. However, a simple shallow-and-wide network outperforms deep models such as DenseNet\cite{Huang2017} with word inputs.
Guo et al.~\cite{Guo2019} study the impact of word embedding and propose to use weighted word embeddings via a multi-channel CNN model. 
Zhang et al.~\cite{zhang2015sensitivity} examine the impact of different word embedding methods and pooling mechanisms, and find that using non-static word2vec and GloVe outperforms one-hot vectors, and that max-pooling consistently outperforms other pooling methods. 

There are other interesting CNN-based models. 
Mou et al.~\cite{mou2015natural} present a tree-based CNN to capture sentence-level semantics.
Pang et al.~\cite{Pang2016} cast text matching as the image recognition task, and use multi-layer CNNs to identify salient n-gram patterns.
Wang et al.~\cite{Wang2017} propose a CNN-based model that combines explicit and implicit representations of short text for TC.
There is also a growing interest in applying CNNs to biomedical text classification~\cite{Karimi2017,Peng2016,Rios2015,Hughes2017}.

\subsection{\textbf{Capsule Neural Networks}}
\label{subsec:capsule}

CNNs classify images or texts by using successive layers of convolutions and pooling. Although pooling operations identify salient features and reduce the computational complexity of convolution operations, they lose information regarding spatial relationships and are likely to mis-classify entities based on their orientation or proportion.

To address the problems of pooling, a new approach is proposed by Hinton et al., called capsule networks (CapsNets)~\cite{hinton2011transforming,sabour2017dynamic}. A capsule is a group of neurons whose activity vector represents different attributes of a specific type of entity such as an object or an object part. The vector's length represents the probability that the entity exists, and the orientation of the vector represents the attributes of the entity. Unlike max-pooling of CNNs, which selects some information and discards the rest, capsules ``route'' each capsule in the lower layer to its best parent capsule in the upper layer, using all the information available in the network up to the final layer for classification. 
Routing can be implemented using different algorithms, such as dynamic routing-by-agreement~\cite{sabour2017dynamic} or the EM algorithm~\cite{sabour2018matrix}. 
 
\begin{figure}[h]
  \centering
  \includegraphics[width=0.35\linewidth]{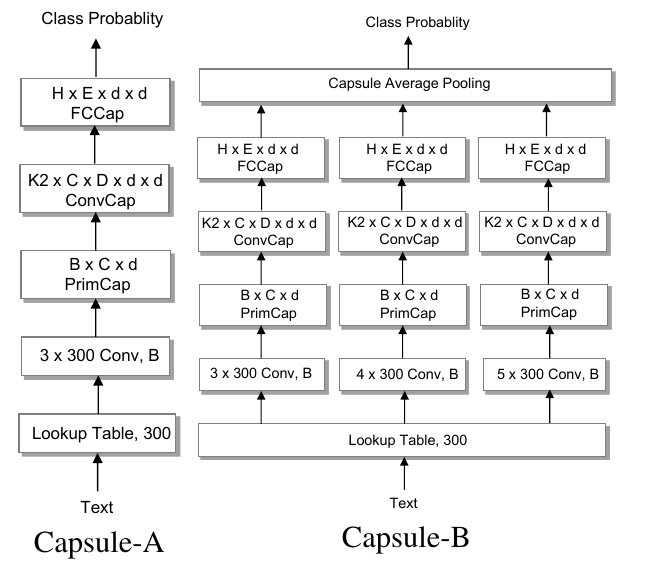}
  \caption{CapsNet A and B for text classification~\cite{zhao2018investigating}.}%
  \label{fig:capsa_capsb}%
\end{figure}

Recently, capsule networks have been applied to TC, where capsules are adapted to represent a sentence or document as a vector. 
\cite{zhao2018investigating,yang2019investigating,zhatow} propose a TC model based on a variant of CapsNets. The model consists of four layers: (1) an n-gram convolutional layer, (2) a capsule layer, (3) a convolutional capsule layer, and (4) a fully connected capsule layer. 
The authors experiment three strategies to stabilize the dynamic routing process to alleviate the disturbance of the noise capsules that contain background information such as stopwords or the words that are unrelated to any document categories.
They also explore two capsule architectures, Capsule-A and Capsule-B as in Fig.~\ref{fig:capsa_capsb}. Capsule-A is similar to the CapsNet in~\cite{sabour2017dynamic}. Capsule-B uses three parallel networks with filters with different window sizes in the n-gram convolutional layer to learn a more comprehensive text representation. CapsNet-B performs better in the experiments. 

The CapsNet-based model proposed by Kim et al.~\cite{kim2020text} uses a similar architecture. The model consists of (1) an input layer that takes a document as a sequence of word embeddings; (2) a convolutional layer that generates feature maps and uses a gated-linear unit to retain spatial information; (3) a convolutional capsule layer to form global features by aggregating local features detected by the convolutional layer; and (4) a text capsule layer to predict class labels. The authors observe that objects can be more freely assembled in texts than in images. For example, a document's semantics can remain the same even if the order of some sentences is changed, unlike the the positions of the eyes and nose on a human face. Thus, they use a static routing schema, which consistently outperforms dynamic routing~\cite{sabour2017dynamic} for TC. 
Aly et al.~\cite{aly2019hierarchical} propose to use CapsNets for Hierarchical Multilabel Classification (HMC), arguing that the CapsNet's capability of encoding child-parent relations makes it a better solution than traditional methods to the HMC task where documents are assigned one or multiple class labels organized in a hierarchical structure. Their model's architecture is similar to the ones in~\cite{zhao2018investigating,yang2019investigating,kim2020text}. 

Ren et al.~\cite{ren2018compositional} propose another variant of CapsNets using a compositional coding mechanism between capsules and a new routing algorithm based on $k$-means clustering. First, the word embeddings are formed using all codeword vectors in codebooks. Then features captured by the lower-level capsules are aggregated in high-level capsules via $k$-means routing.

\subsection{\textbf{Models with Attention Mechanism}}
\label{subsec:attention}

Attention is motivated by how we pay visual attention to different regions of an image or correlate words in one sentence. Attention becomes an increasingly popular concept and useful tool in developing DL models for NLP \cite{bahdanau2014neural,luong2015effective}.
In a nutshell, attention in language models can be interpreted as a vector of importance weights. In order to predict a word in a sentence, we estimate using the attention vector how strongly it is correlated with, or ``attends to'', other words and take the sum of their values weighted by the attention vector as the approximation of the target.

This section reviews some of the most prominent attention models which create new state of the arts on TC tasks, when they are published.

Yang et al.~\cite{yang2016hierarchical} propose a hierarchical attention network for text classification. This model has two distinctive characteristics: (1) a hierarchical structure that mirrors the hierarchical structure of documents, and (2) two levels of attention mechanisms applied at the word and sentence-level, enabling it to attend differentially to more and less important content when constructing the document representation. This model outperforms previous methods by a substantial margin on six TC tasks.
Zhou et al.~\cite{zhou2016attention} extend the hierarchical attention model to cross-lingual sentiment classification. In each language, a LSTM network is used to model the documents. Then, classification is achieved by using a hierarchical attention mechanism, where the sentence-level attention model learns which sentences of a document are more important for determining the overall sentiment. while the word-level attention model learns which words in each sentence are decisive. 

Shen et al.~\cite{shen2018disan} present a directional self-attention network for RNN/CNN-free language understanding, where the attention between elements from input sequence(s) is directional and multi-dimensional. A light-weight neural net is used to learn sentence embedding, solely based on the proposed attention without any RNN/CNN structure. 
Liu et al.~\cite{liu2016learning} present a LSTM model with inner-attention for NLI. This model uses a two-stage process to encode a sentence.
Firstly, average pooling is used over word-level Bi-LSTMs to generate a first stage sentence representation. Secondly, attention mechanism is employed to replace average pooling on the same sentence for better representations. The sentence's first-stage representation is used to attend words appeared in itself. 

\begin{figure}
\centering
\includegraphics[width=0.9\linewidth]{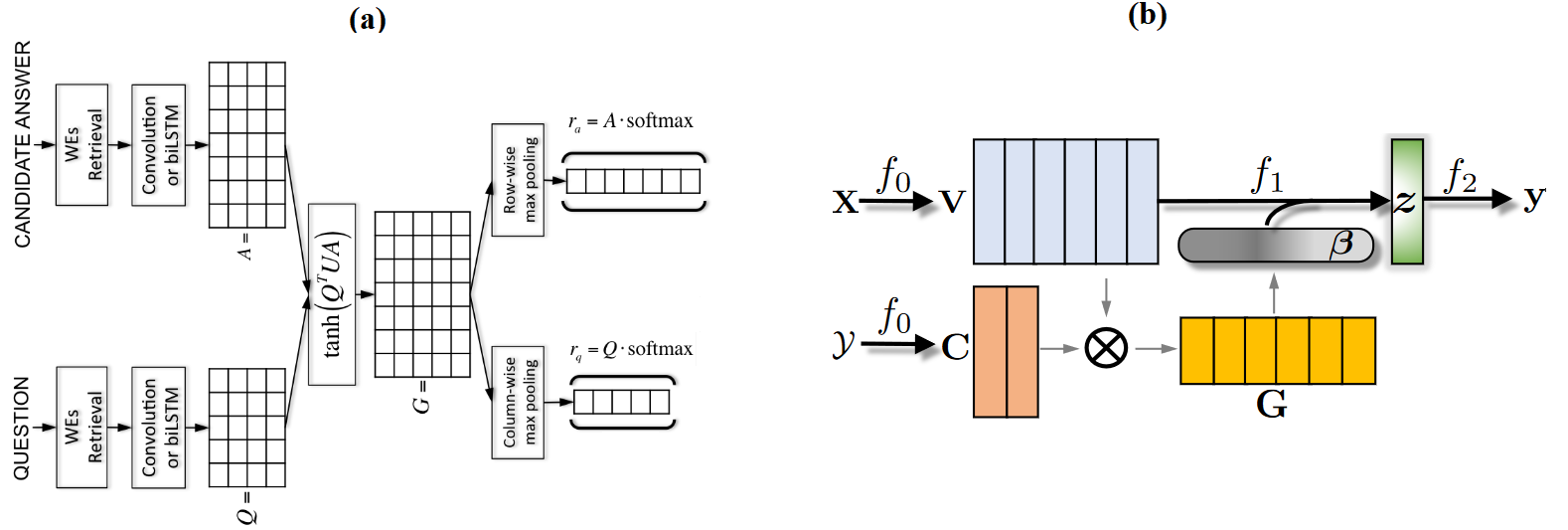}
\caption{(a) The architecture of attentive pooling networks~\cite{santos2016attentive}. (b) The architecture of label-text matching model~\cite{wang2018joint}.}
\label{fig:att}%
\end{figure}

Attention models are widely applied to pair-wise ranking or text matching tasks too.
Santos et al.~\cite{santos2016attentive} propose a two-way attention mechanism, known as Attentive Pooling (AP), for pair-wise ranking. AP enables the pooling layer to be aware of the current input pair (e.g., a question-answer pair), in a way that information from the two input items can directly influence the computation of each other's representations. In addition to learning the representations of the input pair, AP jointly learns a similarity measure over projected segments of the pair, and subsequently derives the corresponding attention vector for each input to guide the pooling. AP is a general framework independent of the underlying representation learning, and can be applied to both CNNs and RNNs, as illustrated in Fig.~\ref{fig:att} (a).
Wang et al.~\cite{wang2018joint} view TC as a label-word matching problem: each label is embedded in the same space with the word vector. The authors introduce an attention framework that measures the compatibility of embeddings between text sequences and labels via cosine similarity, as shown in Fig.~\ref{fig:att} (b).

Kim et al.~\cite{kim2019semantic} propose a semantic sentence matching approach using a densely-connected recurrent and co-attentive network. Similar to DenseNet~\cite{Huang2017}, each layer of this model uses concatenated information of attentive features as well as hidden features of all the preceding recurrent layers. It enables preserving the original and the co-attentive feature information from the bottom-most word embedding layer to the uppermost recurrent layer.
Yin et al.~\cite{yin2016abcnn} present another attention-based CNN model for sentence pair matching. They examine three attention schemes for integrating mutual influence between sentences into CNNs, so that the representation of each sentence takes into consideration its paired sentence. These interdependent sentence pair representations are shown to be more powerful than isolated sentence representations, as validated on multiple classification tasks including answer selection, paraphrase identification, and textual entailment.
Tan et al.~\cite{tan2018multiway} employ multiple attention functions to match sentence pairs under the matching-aggregation framework.
Yang et al.~\cite{yang2016anmm} introduce an attention-based neural matching model for ranking short answer texts. They adopt value-shared weighting scheme instead of position-shared weighting scheme for combining different matching signals and incorporated question term importance learning using question attention network.
This model achieves promising results on the TREC QA dataset.

There are other interesting attention models.
Lin et al.~\cite{lin2017structured} used self-attention to extract interpretable sentence embeddings.
Wang et al.~\cite{wang2018densely} proposed a densely connected CNN with multi-scale feature attention to produce variable n-gram features.
Yamada and Shindo~\cite{yamada2019neural} used neural attentive bag-of-entities models to perform TC using entities in a knowledge base. 
Parikh et al.~\cite{parikh2016decomposable} used attention to decompose a problem into sub-problems that can be solved separately. 
Chen et al.~\cite{chen2018enhancing} explored generalized pooling methods to enhance sentence embedding, and proposed a vector-based multi-head attention model.
Basiri et al.~\cite{basiri2020abcdm} proposed an attention-based bidirectional CNN-RNN deep model for sentiment analysis.

\subsection{\textbf{Memory-Augmented Networks}}
\label{subsec:memory-networks}

While the hidden vectors stored by an attention model during encoding can be viewed as entries of the model's \emph{internal memory}, memory-augmented networks combine neural networks with a form of \emph{external memory}, which the model can read from and write to.

\begin{figure}[h]
  \centering
  {{\includegraphics[width=0.4\linewidth]{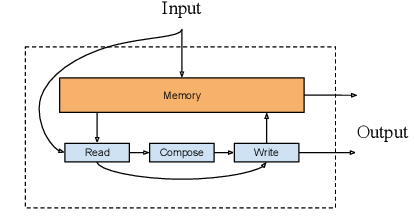} }}%
  \caption{The architecture of NSE~\cite{munkhdalai2017neural}.}%
  \label{fig:nse}%
\end{figure}

Munkhdalai and Yu~\cite{munkhdalai2017neural} present a memory-augmented neural network, called Neural Semantic Encoder (NSE), for TC and QA. NSE is equipped with a variable sized encoding memory that evolves over time and maintains the understanding of input sequences through read, compose and write operations, as shown in Fig.~\ref{fig:nse}.

Weston et al.~\cite{Weston2015} design a memory network for a synthetic QA task, where a series of statements (memory entries) are provided to the model as supporting facts to the question. The model learns to retrieve one entry at a time from memory based on the question and previously retrieved memory. 
Sukhbaatar et al.~\cite{sukhbaatar2015end} extend this work and propose end-to-end memory networks, where memory entries are retrieved in a soft manner with attention mechanism, thus enabling end-to-end training. They show that with multiple rounds (hops), the model is able to retrieve and reason about several supporting facts to answer a specific question.

Kumar et al.~\cite{Kumar2016} propose a Dynamic Memory Metwork (DMN), which processes input sequences and questions, forms episodic memories, and generates relevant answers. Questions trigger an iterative attention process, which allows the model to condition its attention on the inputs and the result of previous iterations. These results are then reasoned over in a hierarchical recurrent sequence model to generate answers. The DMN is trained end-to-end, and obtains state of the art results on QA and POS tagging. 
Xiong et al.~\cite{Xiong2016} present a detailed analysis of the DMN, and improve its memory and input modules.

\subsection{\textbf{Graph Neural Networks}}
\label{subsec:gnn}

Although natural language texts exhibit a sequential order, they also contain internal graph structures, such as syntactic and semantic parse trees, which define the syntactic and semantic relations among words in sentences. 

One of the earliest graph-based models developed for NLP is TextRank~\cite{mihalcea2004textrank}. The authors propose to represent a natural language text as a graph $G(V,E)$, where $V$ denotes a set of nodes and $E$ a set of edges among the nodes. Depending on the applications at hand, nodes can represent text units of various types, e.g., words, collocations, entire sentences, etc. Similarly, edges can be used to represent different types of relations between any nodes, e.g., lexical or semantic relations, contextual overlap, etc. 

Modern Graph Neural Networks (GNNs) are developed by extending DL approaches for graph data, such as the text graphs used by TextRank. Deep neural networks, such as CNNs, RNNs and autoencoders, have been generalized over the last few years to handle the complexity of graph data~\cite{wu2019comprehensive}. 
For example, a 2D convolution of CNNs for image processing is generalized to perform graph convolutions by taking the weighted average of a node's neighborhood information.
Among various types of GNNs, convolutional GNNs, such as Graph Convolutional Networks (GCNs)~\cite{kipf2016semi} and their variants, are the most popular ones because they are effective and convenient to compose with other neural networks, and have achieved state of the art results in many applications. 
GCNs are an efficient variant of CNNs on graphs. 
GCNs stack layers of learned first-order spectral filters followed by a nonlinear activation function to learn graph representations.

A typical application of GNNs in NLP is TC. GNNs utilize the inter-relations of documents or words to infer document labels~\cite{kipf2016semi,hamilton2017inductive,velivckovic2017graph}.
In what follows, we review some variants of GCNs that are developed for TC.

\begin{figure}[h]
\begin{center}
  \includegraphics[page=1,width=0.64\linewidth]{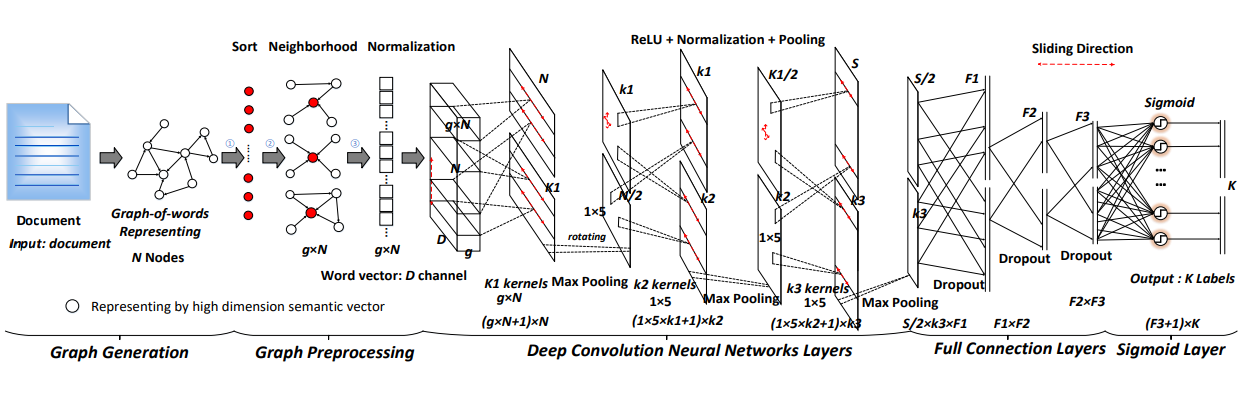}
\end{center}
  \caption{The architecture of GNN used by Peng et al.~\cite{peng2018large}.}
\label{GNN-peng_larg}
\end{figure}

Peng et al.~\cite{peng2018large} propose a graph-CNN based DL model to first convert text to graph-of-words, and then use graph convolution operations to convolve the word graph, as shown in Fig.~\ref{GNN-peng_larg}. 
They show through experiments that the graph-of-words representation of texts has the advantage of capturing non-consecutive and long-distance semantics, and CNN models have the advantage of learning different level of semantics.

In~\cite{peng2019hierarchical}, Peng et al.~propose a TC model based on hierarchical taxonomy-aware and attentional graph capsule CNNs. One unique feature of the model is the use of the hierarchical relations among the class labels, which in previous methods are considered independent. Specifically, to leverage such relations, the authors develop a hierarchical taxonomy embedding method to learn their representations, and define a novel weighted margin loss by incorporating the label representation similarity. 

\begin{figure}[h]
\begin{center}
  \includegraphics[page=1,width=0.42\linewidth]{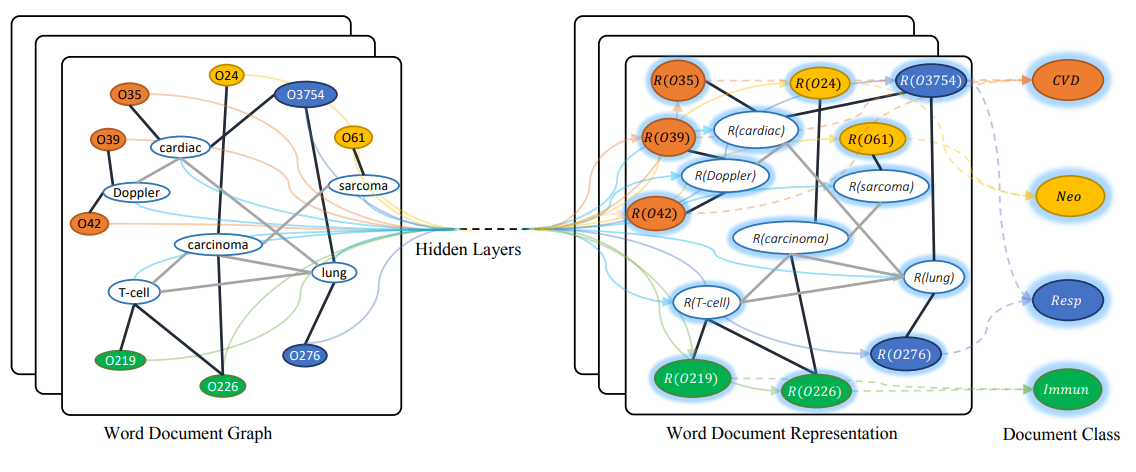}
\end{center}
  \caption{The architecture of GCNN~\cite{yao2019graph}.}
\label{GCNN}
\end{figure}

Yao et al.~\cite{yao2019graph} use a similar Graph CNN (GCNN) model for TC. They build a single text graph for a corpus based on word co-occurrence and document word relations, then learn a Text Graph Convolutional Network (Text GCN) for the corpus, as shown in Fig.~\ref{GCNN}.
The Text GCN is initialized with one-hot representation for word and document, and then jointly learns the embeddings for both words and documents, as supervised by the known class labels for documents.

Building GNNs for a large-scale text corpus is costly. There have been works on reducing the modeling cost by either reducing the model complexity or changing the model training strategy. 
An example of the former is the Simple Graph Convolution (SGC) model proposed in~\cite{wu2019simplifying}, where a deep convolutional GNN is simplified by repeatedly removing the non-linearities between consecutive layers and collapsing the resulting functions (weight matrices) into a single linear transformation. 
An example of the latter is the text-level GNN~\cite{huang2019text}. Instead of building a graph for an entire text corpus, a text-level GNN produces one graph for each text chunk defined by a sliding window on the text corpus so as to reduce the memory consumption during training. 
Some of the other promising GNN based works include,  GraphSage~\cite{hamilton2017inductive}, and contextualized non-local neural networks \cite{liu2019contextualized}.

\subsection{\textbf{Siamese Neural Networks}}
\label{subsec:siamese-networks}

Siamese neural networks (S2Nets)~\cite{BROMLEY1993,Yih2011} and their DNN variants, known as Deep Structured Semantic Models (DSSMs)~\cite{huang2013learning,shen2014latent}, are designed for text matching. 
The task is fundamental to many NLP applications, such as query-document ranking and answer selection in extractive QA. These tasks can be viewed as special cases of TC. For example, in question-document ranking, we want to classify a document as relevant or irrelevant to a given query. 

\begin{figure}[t]
\centering 
\includegraphics[width=0.22\linewidth]{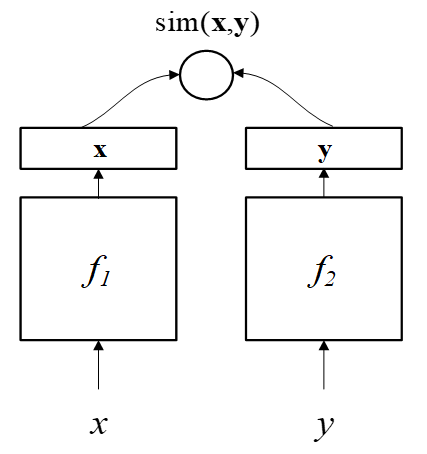}
\vspace{-2mm}
\caption{The architecture of a DSSM, illustrated in ~\cite{gao2019neural}} 
\label{fig:dssm} 
\vspace{-2mm}
\end{figure}

As illustrated in Fig.~\ref{fig:dssm}, a DSSM (or a S2Net) consists of a pair of DNNs, $f_1$ and $f_2$, which map inputs $x$ and $y$ into corresponding vectors in a common low-dimensional semantic space~\cite{gao2019neural}. 
Then the similarity of $x$ and $y$ is measured by the cosine distance of the two vectors. 
While S2Nets assume that $f_1$ and $f_2$ share the same architecture and even the same parameters, in DSSMs, 
$f_1$ and $f_2$ can be of different architectures depending on $x$ and $y$. For example, to compute the similarity of an image-text pair, $f_1$ can be a deep CNN and $f_2$ an RNN or MLP.
These models can be applied to a wide range of NLP tasks depending on the definition of $(x, y)$. 
For example, $(x, y)$ could be a query-document pair for query-document ranking \citep{shen2014latent,Severyn2015}, or a question-answer pair in QA \citep{Das2016,Tan2016}.

The model parameters $\theta$ are often optimized using a pair-wise rank loss. 
Take document ranking as an example. 
Consider a query $x$ and two candidate documents $y^+$ and $y^-$, where $y^+$ is relevant to $x$ and $y^-$ is not. 
Let $\text{sim}_{\theta} (x,y)$ be the cosine similarity of $x$ and $y$ in the semantic space parameterized by $\theta$. The training objective is to minimize the margin-based loss as
\begin{equation}
\label{eq:dssm_loss}
L(\theta)= \left[ \gamma + \text{sim}_{\theta} (x,y^-)-\text{sim}_{\theta} (x, y^+) \right]_+, 
\end{equation}
where $[x]_+ := \max(0,x)$ and $\gamma$ is the margin hyperparameter.

\begin{figure}[h]
\begin{center}
  \includegraphics[page=1,width=0.31\linewidth]{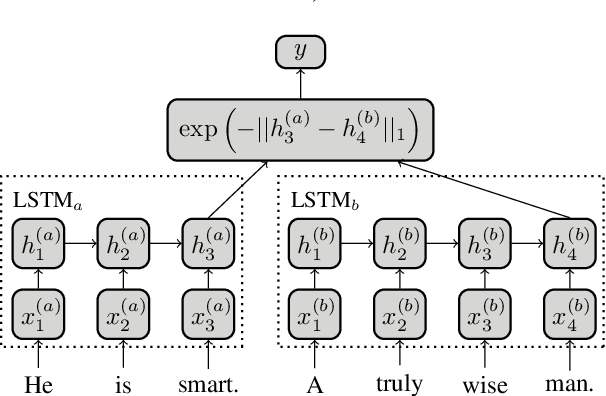}
\end{center}
  \caption{The architecture of the Siamese model proposed by Mueller et al.~\cite{Mueller2016}.}
\label{Mueller-SiameseRNN}
\end{figure}

Since texts exhibit a sequential order, it is natural to implement $f_1$ and $f_2$ using RNNs or LSTMs to measure the semantic similarity between texts.
Fig.~\ref{Mueller-SiameseRNN} shows the architecture of the siamese model proposed in~\cite{Mueller2016}, where the two networks use the same LSTM model. 
Neculoiu et al.~\cite{Neculoiu2016} present a similar model that uses character-level Bi-LSTMs for $f_1$ and $f_2$, and the cosine function to calculate the similarity. 
Liu et al.~\cite{liu2016modelling} model the interaction of a sentence pair with two coupled-LSTMs. 
In addition to RNNs, BOW models and CNNs are also used in S2Nets to represent sentences.
For example, He et al.~\cite{He2015} propose a S2Net that uses CNNs to model multi-perspective sentence similarity.
Renter et al.~\cite{Renter2016} propose a Siamese CBOW model which forms a sentence vector representation by averaging the word embeddings of the sentence, and calculates the sentence similarity as cosine similarity between sentence vectors.
As BERT becomes the new state of the art sentence embedding model, there have been attempts to building BERT-based S2Nets, such as SBERT~\cite{Reimers2019} and TwinBERT~\cite{lu2020twinbert}.

\begin{figure}
  \centering
    \includegraphics[page=1,width=0.31\linewidth]{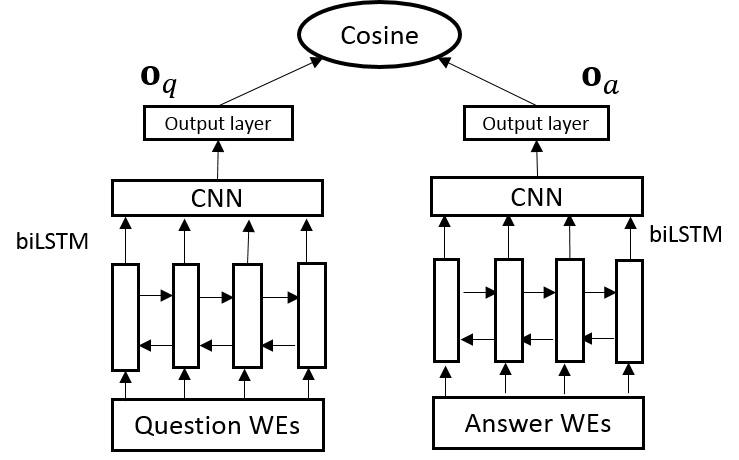}
    \includegraphics[page=1,width=0.31\linewidth]{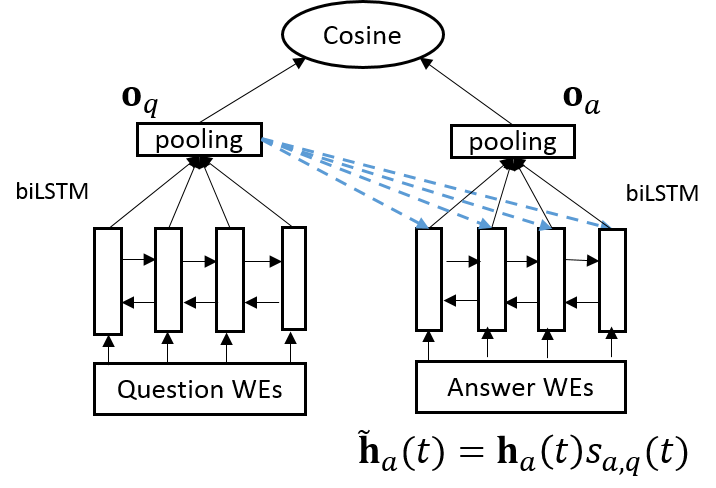}
  \caption{The architectures of the Siamese models studied in~\cite{Tan2016}.}
\label{Tan-Siamese}
\end{figure}

S2Nets and DSSMs have been widely used for QA. 
Das et al.~\cite{Das2016} propose a Siamese CNN for QA (SCQA) to measure the semantic similarity between a question and its (candidate) answers. To reduce the computational complexity, SCQA uses character-level representations of question-answer pairs. The parameters of SCQA is trained to maximize the semantic similarities between a question and its relevant answers, as Equation~\ref{eq:dssm_loss}, where $x$ is a question and $y$ its candidate answer.
Tan et al.~\cite{Tan2016} present a series of siamese neural networks for answer selection. As shown in Fig.~\ref{Tan-Siamese}, these are hybrid models that process text using convolutional, recurrent, and attention neural networks. 
Other siamese neural networks developed for QA include LSTM-based models for non-factoid answer selection~\cite{tan2015lstm}, Hyperbolic representation learning \cite{tay2018hyperbolic}, and QA using a deep similarity neural network \cite{minaee2017automatic}.

\subsection{\textbf{Hybrid Models}}
\label{subsec:hybrid-models}

\begin{figure}[h]
  \centering
  \includegraphics[page=1,width=0.8\linewidth]{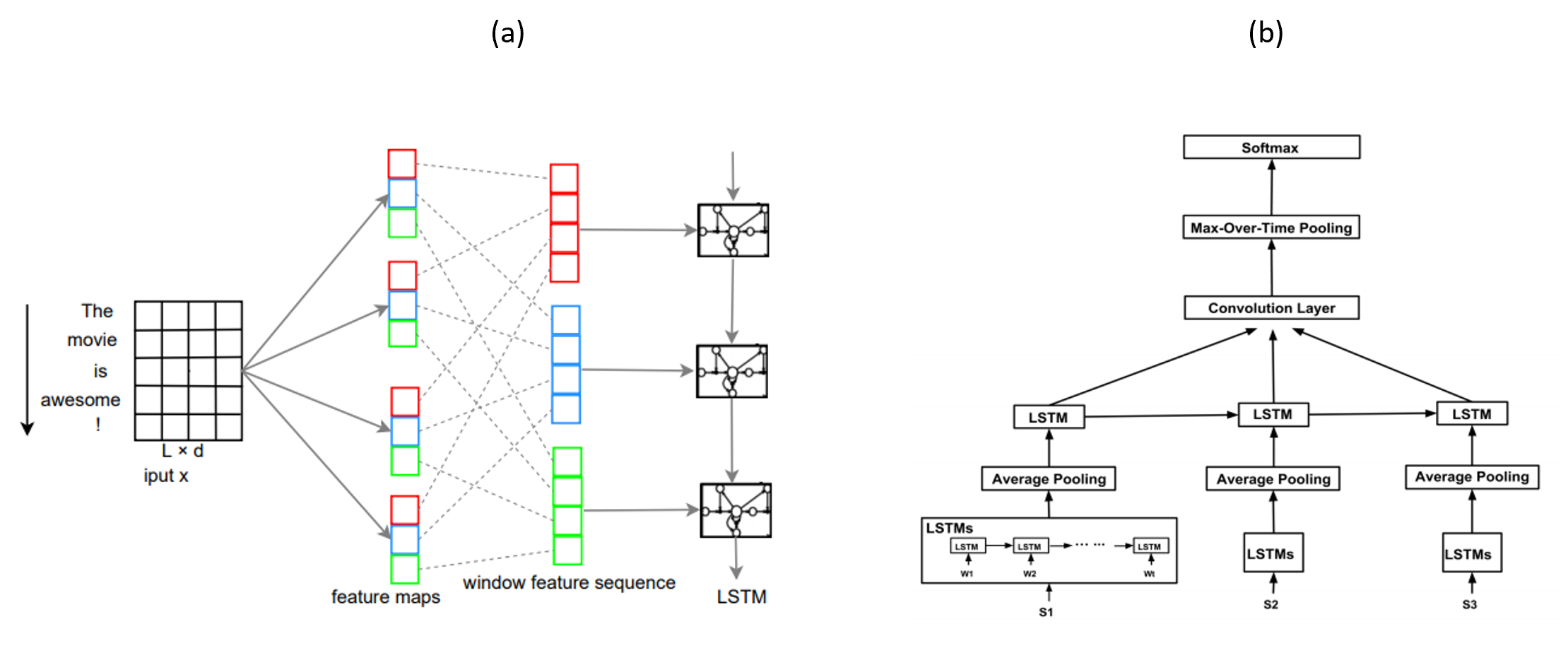}
  \caption{ (a) The architecture of C-LSTM~\cite{zhou2015c}. (b) The architecture of DSCNN for document modeling~\cite{Zhang2016}.}
  \label{fig:lstm-cnn}
\end{figure}

Many Hybrid models have been developed to combine LSTM and CNN architectures to capture local and global features of sentences and documents.
Zhu et al.~\cite{zhou2015c} propose a Convolutional LSTM (C-LSTM) network. As illustrated in Fig.~\ref{fig:lstm-cnn} (a), C-LSTM utilizes a CNN to extract a sequence of higher-level phrase (n-gram) representations, which are fed to a LSTM network to obtain the sentence representation.
Similarly, Zhang et al.~\cite{Zhang2016} propose a Dependency Sensitive CNN (DSCNN) for document modeling. As illustrated in Fig.~\ref{fig:lstm-cnn} (b), the DSCNN is a hierarchical model, where LSTM learns the sentence vectors which are fed to the convolution and max-pooling layers to generate the document representation.

Chen et al.~\cite{cheens} perform multi-label TC through a CNN-RNN model that is able to capture both global and local textual semantics and, hence, to model high-order label correlations while having a tractable computational complexity. 
Tang et al.~\cite{tang2015document} use a CNN to learn sentence representations, and a gated RNN to learn a document representation that encodes the intrinsic relations between sentences.
Xiao et al.~\cite{xiao2016efficient} view a document as a sequence of characters, instead of words, and propose to use both character-based convolution and recurrent layers for document encoding. This model achieved comparable performances with much less parameters, compared with word-level models.
The Recurrent CNN~\cite{lai2015recurrent} applies a recurrent structure to capture long-range contextual dependence for learning word representations. To reduce the noise, max-pooling is employed to automatically select only the salient words that are crucial to the text classification task.

Chen et al.~\cite{CHEN2017221} propose a divide-and-conquer approach to sentiment analysis via sentence type classification, motivated by the observation that different types of sentences express sentiment in very different ways. The authors first apply a Bi-LSTM model to classify opinionated sentences into three types. Each group of sentences is then fed to a one-dimensional CNN separately for sentiment classification.

In~\cite{kowsari2017hdltex}, Kowsari et al.~propose a Hierarchical Deep Learning approach for Text classification (HDLTex). HDLTex employs stacks of hybrid DL model architectures, including MLP, RNN and CNN, to provide specialized understanding at each level of the document hierarchy. 

Liu~\cite{liu2017stochastic} propose a robust Stochastic Answer Network (SAN) for multi-step reasoning in machine reading comprehension. 
SAN combines neural networks of different types, including memory networks, Transforms, Bi-LSTM, attention and CNN. 
The Bi-LSTM component obtains the context representations for questions and passages. Its attention mechanism derives a question-aware passage representation.
Then, another LSTM is used to generate a working memory for the passage.
Finally, a Gated Recurrent Unit based answer module outputs predictions. 

\begin{figure}[h]
  \centering
   \includegraphics[page=1,width=0.61\linewidth]{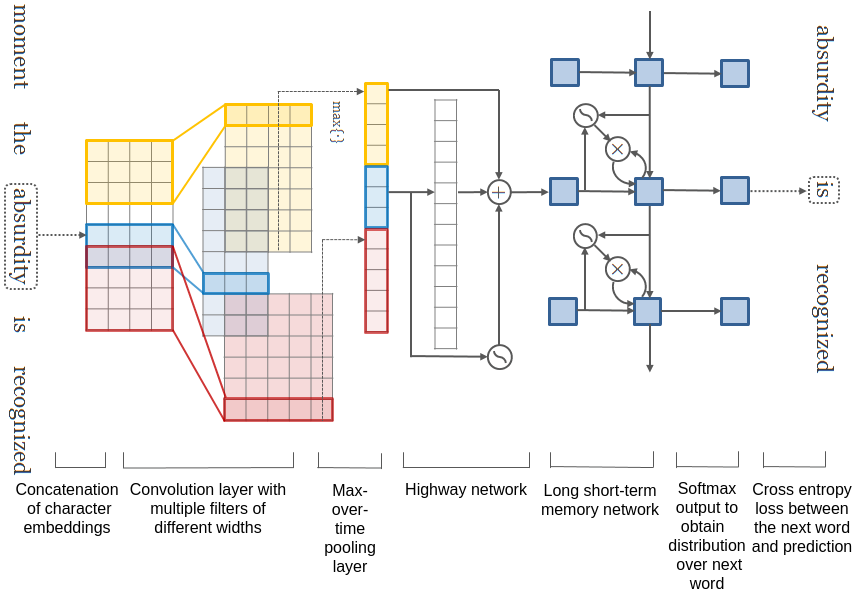}
  \caption{The architecture of the highway network with CNN and LSTM~\cite{Kim2016}.}
  \label{fig:hybrid:3}
\end{figure}

Several studies have been focused on combining highway networks with RNNs and CNNs. 
In typical multi-layer neural networks, information flows layer by layer. Gradient-based training of a DNN becomes more difficult with increasing depth.  
Highway networks~\cite{Srivastava2015} are designed to ease training of very deep neural networks. They allow unimpeded information flow across several layers on \emph{information highways}, similar to the shortcut connections in ResNet~\cite{he2016deep}. 
Kim et al.~\cite{Kim2016} employ a highway network with CNN and LSTM over characters for language modeling. As illustrated in Fig.~\ref{fig:hybrid:3}, the first layer performs a lookup of character embeddings, then convolution and max-pooling operations are applied to obtain a fixed-dimensional representation of the word, which is given to the highway network. The highway network's output is used as the input to a multi-layer LSTM. Finally, an affine transformation followed by a softmax is applied to the hidden representation of the LSTM to obtain the distribution over the next word. 
Other highway-based hybrid models include 
recurrent highway networks~\cite{Zilly2017}, 
and RNN with highway~\cite{wen2016learning}.

\subsection{\textbf{Transformers and Pre-Trained Language Models}}
\label{subsec:transformers}

One of the computational bottlenecks suffered by RNNs is the sequential processing of text. 
Although CNNs are less sequential than RNNs, the computational cost to capture relationships between words in a sentence also grows with the increasing length of the sentence, similar to RNNs.
Transformers~\cite{vaswani2017attention} overcome this limitation by applying self-attention to compute in parallel for every word in a sentence or document an ``attention score'' to model the influence each word has on another
\footnote{Strictly speaking, Transformer is an instance of hybrid models (\ref{subsec:hybrid-models}), since each Transformer layer is a composite structure consisting of a feed-forward layer and a multi-head attention layer.}.
Due to this feature, Transformers allow for much more parallelization than CNNs and RNNs, which makes it possible to efficiently train very big models on large amounts of data on GPUs.

Since 2018 we have seen the rise of a set of large-scale Transformer-based Pre-trained Language Models (PLMs).
Compared to earlier contextualized embedding models based on CNNs~\cite{collobert2011natural} or LSTMs~\cite{peters2018deep}, Transformer-based PLMs use much deeper network architectures (e.g., 48-layer Transformers~\cite{radford2019language}), and are \emph{pre-trained} on much larger amounts of text corpora to learn contextual text representations by predicting words conditioned on their context. These PLMs are \emph{fine-tuned} using task-specific labels, and have created new state of the art in many downstream NLP tasks, including TC. Although pre-training is unsupervised (or self-supervised), fine-tuning is supervised learning. 
A recent survey of Qiu et al. \cite{qiu2020pre} categories popular PLMs by their representation types, model architectures, pre-training tasks, and downstream tasks. 

\begin{figure}[h]
  \centering
  \includegraphics[page=1,width=0.72\linewidth]{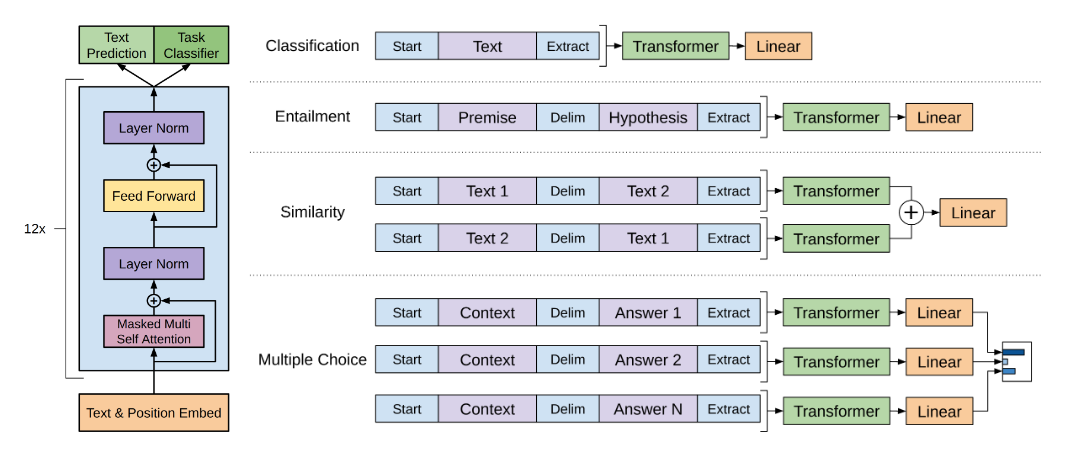}
  \caption{The architecture of OpenGPT-1~\cite{radford2019language}}
  \label{fig:opengpt1}
\end{figure}

PLMs can be grouped into two categories, autoregressive and autoencoding PLMs.
One of the earliest autoregressive PLMs is OpenGPT~\cite{radford2018improving,radford2019language}, a unidirectional model which predicts a text sequence word by word from left to right (or right to left), with each word prediction depending on previous predictions. 
Fig.~\ref{fig:opengpt1} shows the architecture of OpenGPT. It consists of 12 layers of Transformer blocks, each consisting of a masked multi-head attention module, followed by a layer normalization and a position-wise feed forward layer. OpenGPT can be adapted to downstream tasks such as TC by adding task-specific linear classifiers and fine-tuning using task-specific labels.

One of the most widely used autoencoding PLMs is BERT~\cite{devlin2018bert}. 
Unlike OpenGPT which predicts words based on previous predictions, BERT is trained using the masked language modeling (MLM) task that randomly masks some tokens in a text sequence, and then independently recovers the masked tokens by conditioning on the encoding vectors obtained by a bidirectional Transformer. 
There have been numerous works on improving BERT. 
RoBERTa~\cite{liu2019roberta} is more robust than BERT, and is trained using much more training data. 
ALBERT~\cite{lan2019albert} lowers the memory consumption and increases the training speed of BERT. 
DistillBERT~\cite{sanh2019distilbert} utilizes knowledge distillation during pre-training to reduce the size of BERT by 40\% while retaining 99\% of its original capabilities and making the inference 60\% faster. 
SpanBERT~\cite{joshi2019spanbert} extends BERT to better represent and predict text spans. 
Electra~\cite{clark2020electra} uses a more sample-efficient pre-training task than MLM, called replaced token detection. Instead of masking the input, it corrupts it by replacing some tokens with plausible alternatives sampled from a small generator network.
ERNIE~\cite{sun2019ernie,sun2020ernie} incorporates domain knowledge from external knowledge bases, such as named entities, for model pre-training. 
ALUM~\cite{liu2020adversarial} introduces an adversarial loss for model pretraining that improves the model's generalization to new tasks and robustness to adversarial attacks. 
BERT and its variants have been fine-tuned for various NLP tasks, including QA~\cite{garg2019tanda}, TC~\cite{sun2019fine}, and NLI~\cite{zhang2019semantics, liu2019multi}.

\begin{figure}[h]
  \centering
  \includegraphics[page=1,width=0.7\linewidth]{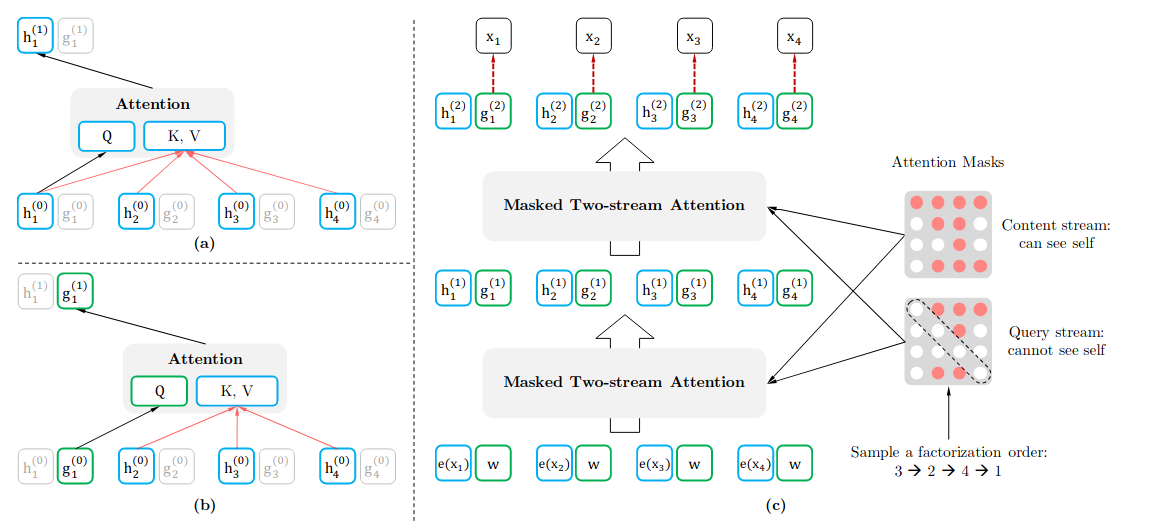}
  \caption{The architecture of XLNet~\cite{yang2019xlnet}: a) Content stream attention, b) Query stream attention, c) Overview of the permutation language modeling training with two- stream attention.}
  \label{fig:xlnet}
\end{figure}

There have been attempts to combine the strengths of autoregressive and autoencoding PLMs.
XLNet~\cite{yang2019xlnet} integrates the idea of autoregressive models like OpenGPT and bi-directional context modeling of BERT.
XLNet makes use of a \emph{permutation operation} during pre-training that allows context to include tokens from both left and right, making it a generalized order-aware autoregressive language model. The permutation is achieved by using a special attention mask in Transformers.
XLNet also introduces a two-stream self-attention schema to allow position-aware word prediction. This is motivated by the observation that word distributions vary greatly depending on word positions. For example, the beginning of a sentence has a considerably different distribution from other positions in the sentence.
As show in Fig.~\ref{fig:xlnet}, to predict the word token in position 1 in a permutation 3-2-4-1, a content stream is formed by including the positional embeddings and token embeddings of all previous words (3, 2, 4), then a query stream is formed by including the content stream and the positional embedding of the word to be predicted (word in position 1), and finally the model makes the prediction based on information from the query stream.

\begin{figure}[h]
  \centering
  \includegraphics[page=1,width=0.66\linewidth]{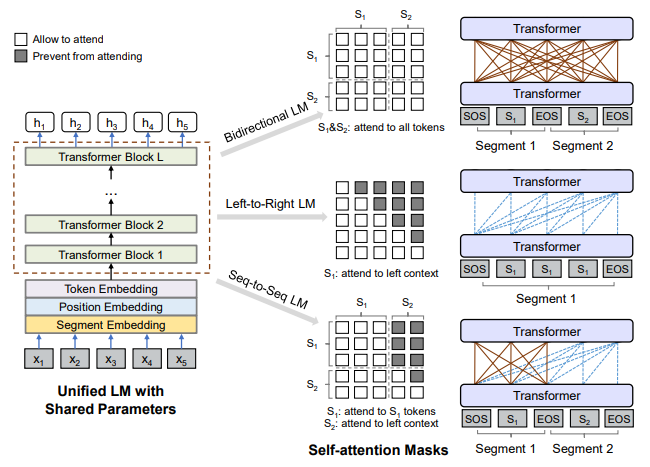}
  \caption{Overview of UniLM pre-training~\cite{dong2019unified}. The model parameters are shared across the language modeling objectives i.e., bidirectional, unidirectional, and sequence-to-sequence language modeling. Different self-attention masks are used to control the access to context for each word token.}
  \label{fig:unilm}
\end{figure}

As mentioned earlier, OpenGPT uses a left-to-right Transformer to learn text representation for natural language generation, while BERT uses a bidirectional transformer for natural language understanding. 
The Unified language Model (UniLM)~\cite{dong2019unified} is designed to tackle both natural language understanding and generation tasks. 
UniLM is pre-trained using three types of language modeling tasks: unidirectional, bidirectional, and sequence-to-sequence prediction. 
The unified modeling is achieved by employing a shared Transformer network and utilizing specific self-attention masks to control what context the prediction conditions on, as shown in Fig.~\ref{fig:unilm}. 
The second version of UniLM~\cite{bao2020unilmv2} is reported to achieve new state of the art on a wide range of natural language understanding and generation tasks, significantly outperforming previous PLMs, including OpenGPT-2, XLNet, BERT and its variants.

Raffel et al.~\cite{raffel2019exploring} present a unified Transformer-based framework that converts many NLP problems into a text-to-text format. They also conduct a systematic study to compare pre-training objectives, architectures, unlabeled datasets, fine-tuning approaches, and other factors on dozens of language understanding tasks.

\subsection{\textbf{Beyond Supervised Learning}}
\label{subsec:beyond-supervised-learning}

\paragraph{\textbf{Unsupervised Learning using Autoencoders}}

Similar to word embeddings, distributed representations for sentences can also be learned in an unsupervised fashion.
by optimizing some auxiliary objectives, such as the reconstruction loss of an autoencoder~\cite{rumelhart1985learning}. The result of such unsupervised learning are sentence encoders, which can map sentences with similar semantic and syntactic properties to similar fixed-size vector representations.
The Transformer-based PLMs described in Section \ref{subsec:transformers} are also unsupervised models that can be used as sentence encoders. This section discusses unsupervised models based on auto-encoders and their variants.

Kiros et al.~\cite{kiros2015skip} propose the Skip-Thought model for unsupervised learning of a generic, sentence encoder. An encoder-decoder model is trained to reconstruct the surrounding sentences of an encoded sentence. 
Dai and Le~\cite{Dai2015} investigate the use of a sequence autoencoder, which reads the input sequence into a vector and predicts the input again, for sentence encoding. They show that pre-training sentence encoders on a large unsupervised corpus yields better accuracy than only pre-training word embeddings.
Zhang et al.~\cite{Zhang2019learning} propose a mean-max attention autoencoder, which uses the multi-head self-attention mechanism to reconstruct the input sequence. A mean-max strategy is used in encoding, where both mean and max pooling operations over the hidden vectors are applied to capture diverse information of the input.

\begin{figure}
\centering
    \includegraphics[width=0.7\linewidth]{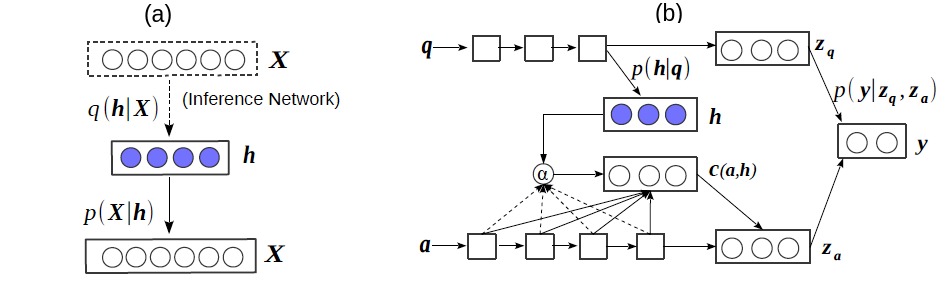} 
    \includegraphics[width=0.55\linewidth]{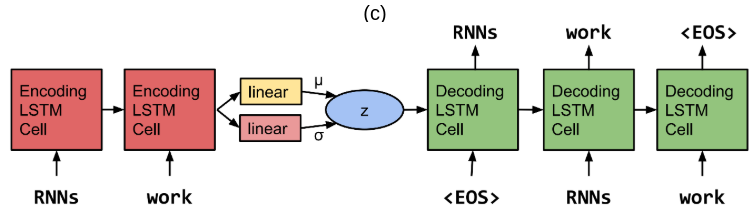} 
\caption{(a) The neural variational document model for document modeling~\cite{miao2016neural}. (b) The neural answer selection model for QA~\cite{miao2016neural}. (c) The RNN-based variational autoencoder language model~\cite{Bowman2016}.}%
    \label{fig:vae}%
\end{figure}

While autoencoders learn a compressed representation of input, Variational AutoEncoders (VAEs)~\cite{Kingma2014,rezende2014stochastic} learn a distribution representing the data, and can be viewed as a regularized version of the autoencoder~\cite{goodfellow2016deep}. Since a VAE learns to model the data, we can easily sample from the distribution to generate new samples (e.g., new sentences). 
Miao et al.~\cite{miao2016neural} extend the VAE framework to text, and propose a Neural Variational Document Model (NVDM) for document modeling and a Neural Answer Selection Model (NASM) for QA.
As shown in Fig.~\ref{fig:vae} (a), the NVDM uses an MLP encoder to map a document to a continuous semantic representation.
As shown in Fig.~\ref{fig:vae} (b), the NASM uses LSTM and a latent stochastic attention mechanism to model the semantics of question-answer pairs and predicts their relatedness. The attention model focuses on the phrases of an answer that are strongly connected to the question semantics and is modeled by a latent distribution, allowing the model to deal with the ambiguity inherent in the task.
Bowman et al.~\cite{Bowman2016} propose an RNN-based VAE language model, as shown in Fig.~\ref{fig:vae} (c). This model incorporates distributed latent representations of entire sentences, allowing to explicitly model holistic properties of sentences such as style, topic, and high-level syntactic features. 
Gururangan et al.~\cite{gururangan2019variational} pre-train a document model as a VAE on in-domain, unlabeled data and use its internal states as features for text classification.
In general, data augmentation using VAE or other models~\cite{Meng2018WeaklySupervisedNT,Chen2020MixTextLI} is  widely used for semi-supervised or weakly supervised TC.

\paragraph{\textbf{Adversarial Training}}

Adversarial training~\cite{goodfellow2014explaining} is a regularization method for improving the generalization of a classifier. It does so by improving model's robustness to adversarial examples, which are created by making small perturbations to the input. Adversarial training requires the use of labels, and is applied to supervised learning.
Virtual adversarial training~\cite{miyato2016distributional} extend adversarial training to semi-supervised learning. This is done by regularizing a model so that given an example, the model produces the same output distribution as it produces on an adversarial perturbation of that example. 
Miyato et al.~\cite{miyato2016adversarial} extend adversarial and virtual adversarial training to supervised and semi-supervised TC tasks by applying perturbations to the word embeddings in an RNN rather than the original input itself.
Sachel et al.~\cite{sachan2019revisiting} study LSTM models for semi-supervised TC. They find that using a mixed objective function that combines cross-entropy, adversarial, and virtual adversarial losses for both labeled and unlabeled data, leads to a significant improvement over supervised learning approaches.
Liu et al.~\cite{liu2017adversarial} extend adversarial training to the multi-task learning framework for TC~\cite{liu2015multi}, aiming to alleviate the task-independent (shared) and task-dependent (private) latent feature spaces from interfering with each other.

\begin{figure}[h]
\begin{center}
  \includegraphics[page=1,width=0.9\linewidth]{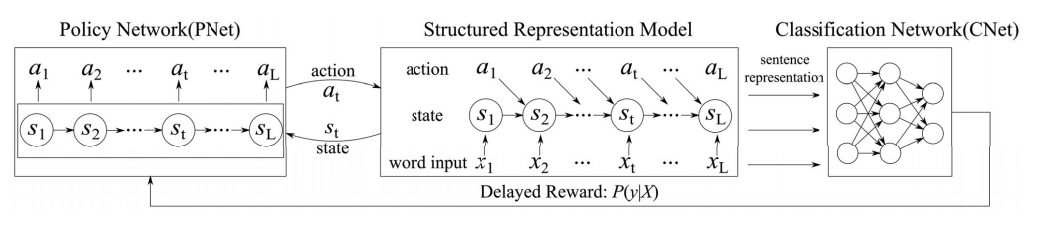}
\end{center}
  \caption{The RL-based method of learning structured representations for text classification~\cite{zhang2018learning}. The policy network samples an action at each state. The structured representation model updates the state and outputs the final sentence representation to the classification network at the end of the episode. The text classification loss is used as a (negative) reward to train the policy.}
\label{rl_zhang}
\end{figure}

\paragraph{\textbf{Reinforcement Learning}}
Reinforcement learning (RL)~\cite{sutton2018reinforcement} is a method of training an agent to perform discrete actions according to a policy, which is trained to maximize a reward. 
Shen et al.~\cite{shen2018reinforced} use a hard attention model to select a subset of critical word tokens of an input sequence for TC. The hard attention model can be viewed as an agent that takes actions of whether to select a token or not. After going through the entire text sequence, it receives a classification loss, which can be used as the reward to train the agent.
Liu et al.~\cite{liu2020finding} propose a neural agent that models TC as a sequential decision process. Inspired by the cognitive process of human text reading, the agent scans a piece of text sequentially and makes classification decisions at the time it wishes. Both the classification result and when to make the classification are part of the decision process, controlled by a policy trained with RL.
Shen et al.~\cite{shen2017reasonet} present a multi-step Reasoning Network (ReasoNet) for machine reading comprehension. ReasoNets tasks multiple steps to reason over the relation among queries, documents, and answers. Instead of using a fixed number of steps during inference, ReasoNets introduce a termination state to relax this constraint on the reasoning steps. With the use of RL, ReasoNets can dynamically determine whether to continue the comprehension process after digesting intermediate results, or to terminate reading when it concludes that existing information is adequate to produce an answer.
Li et al.~\cite{liigen} combine RL, GANs, and RNNs to build a new model, termed Category Sentence Generative Adversarial Network (CS-GAN), which is able to generate category sentences that enlarge the original dataset and to improve its generalization capability during supervised training. 
Zhang et al.~\cite{zhang2018learning} propose a RL-based method of learning structured representations for text classification. They propose two LSTM-based models. The first one selects only important, task-relevant words in the input text. The other one discovers phrase structures of sentences. Structure discovery using these two models is formulated as a sequential decision process guided by a policy network, which decides at each step which model to use, as illustrated in Fig.~\ref{rl_zhang}. The policy network is optimized using policy gradient.

\begin{figure*}[h]
\begin{center}
  \includegraphics[page=1,width=0.90\linewidth]{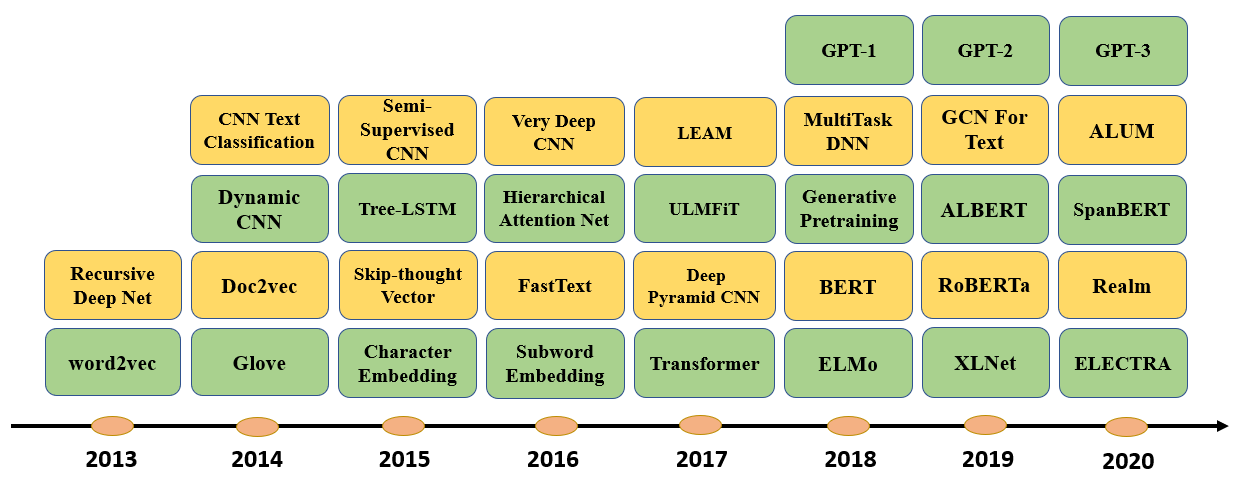}
\end{center}
  \caption{Some of the most prominent deep learning models for text embedding and classification published from 2013 to 2020.}
\label{timeline}
\end{figure*}

As a summary of this section, Figure~\ref{timeline} illustrates the timeline of some of the most popular DL-based models for TC since 2013. 

\section{\textbf{How to Choose the Best Neural Network Model for My Task}}
\label{sec:choose}
The answer to ``what the best neural network architecture is for TC?'' varies greatly depending on the nature of the target task and domain, the availability of in-domain labels, the latency and capacity constraints of the applications, and so on. 
Although it has no doubt that developing a text classifier is a trial-and-error process, 
by analyzing recent results on public benchmarks (e.g., GLUE \cite{wang2018glue}), we propose the following recipe to make the process easier. The recipe consists of five steps:
\begin{enumerate}
    \item \textbf{PLM Selection.} As will be shown in Section \ref{sec:performance}, using PLMs leads to significant improvements across all popular text classification tasks, and autoencoding PLMs (e.g., BERT or RoBERTa) often work better than autoregressive PLMs (e.g., OpenAI GPT). 
    Hugging Face\footnote{https://huggingface.co/} maintains a rich repository of PLMs developed for various tasks and settings.
    \item \textbf{Domain adaptation.} Most PLMs are trained on general-domain text corpora (e.g., Web). If the target domain is dramatically different from general domain, we might consider adapting the PLM using in-domain data by continual pre-training the selected general-domain PLM. For domains with abundant unlabeled text, such as biomedicine, pretraining language models from scratch might also be a good choice \cite{gu2020domain}.
    \item \textbf{Task-specific model design.} Given input text, the PLM produces a sequence of vectors in the contextual representation. Then, one or more task-specific layers are added on the top to generate the final output for the target task. The choice of the architecture of task-specific layers depends on the nature of the task, e.g., the linguistic structure of text needs to be captured.  As described in Section \ref{sec:Deep_text},  feed-forward neural networks view text as a bag of words, RNNs can capture word orders, CNNs are good at recognizing patterns such as key phrases, attention mechanisms are effective to identify correlated words in text, Siamese NNs are used for text matching tasks, and GNNs can be a good choice if graph structures of natural language (e.g., parse trees) are useful for the target task.
    \item \textbf{Task-specific fine-tuning.} Depending on the availability of in-domain labels, the task-specific layers can be either trained alone with the PLM fixed or trained together with the PLM. If multiple similar text classifiers need to be built (e.g., news classifiers for different domains), multi-task fine-tuning \cite{liu2019multi} is a good choice to leverage labeled data of similar domains. 
    \item \textbf{Model compression.} PLMs are expensive to serve. They often need to be compressed via e.g., knowledge distillation \cite{mukherjee2020xtremedistil,tang2019distilling} to meet the latency and capacity constraints in real-world applications. 
\end{enumerate}

\section{\textbf{Text Classification Datasets}}
\label{sec:datasets} 

This section describes the datasets that are widely used for TC research. 
We group these datasets, based on their main target applications, into such categories as sentiment analysis, news categorization, topic classification, QA, and NLI.

\subsection{\textbf{Sentiment Analysis Datasets}}

\paragraph{\textbf{Yelp}} 
Yelp~\cite{yelp} dataset contains the data for two sentiment classification tasks. One is to detect fine-grained sentiment labels and is called Yelp-5. The other predicts the negative and positive sentiments, and is known as Yelp Review Polarity or Yelp-2. 
Yelp-5 has 650,000 training samples and 50,000 test samples for each class, and Yelp-2 includes 560,000 training samples and 38,000 test samples for negative and positive classes. 

\paragraph{\textbf{IMDb}}

The IMDB dataset~\cite{imdb} is developed for the task of binary sentiment classification of movie reviews. IMDB consists of equal number of positive and negative reviews. It is evenly divided between training and test sets with 25,000 reviews for each. 

\paragraph{\textbf{Movie Review}}

The Movie Review (MR) dataset~\cite{movie_review} is a collection of movie reviews developed for the task of detecting the sentiment associated with a particular review and determining whether it is negative or positive. 
It includes 10,662 sentences with even numbers of negative and positive samples. 
10-fold cross validation with random split is usually used for testing on this dataset.

\paragraph{\textbf{SST}} 
The Stanford Sentiment Treebank (SST) dataset ~\cite{socher2013recursive} is an extended version of MR. 
Two versions are available, one with fine-grained labels (five-class) and the other binary labels, referred to as SST-1 and SST-2, respectively.
SST-1 consists of 11,855 movie reviews which are divided into 8,544 training samples, 1,101 development samples, and  2,210 test samples. 
SST-2 is partitioned into three sets with the sizes of 6,920, 872 and 1,821 as training, development and test sets, respectively.

\paragraph{\textbf{MPQA}}

The Multi-Perspective Question Answering (MPQA) dataset~\cite{deng2015mpqa} is an opinion corpus with two class labels. MPQA consists of 10,606 sentences extracted from news articles related to a wide variety of news sources. This is an imbalanced dataset with 3,311 positive documents and 7,293 negative documents.

\paragraph{\textbf{Amazon}}
This is a popular corpus of product reviews collected from the Amazon website~\cite{amazon}. 
It contains labels for both binary classification and multi-class (5-class) classification. 
The Amazon binary classification dataset consists of 3,600,000 and 400,000 reviews for training and test, respectively. 
The Amazon 5-class classification dataset (Amazon-5) consists of 3,000,000 and 650,000 reviews for training and test, respectively.



\subsection{\textbf{News Classification Datasets}}

\paragraph{\textbf{AG News}}

The AG News dataset~\cite{zhang2015character} is a collection of news articles collected from more than 2,000 news sources by ComeToMyHead, an academic news search engine. This dataset includes 120,000 training samples and 7,600 test samples. 
Each sample is a short text with a four-class label.

\paragraph{\textbf{20 Newsgroups}}


The 20 Newsgroups dataset~\cite{news20} is a collection of newsgroup documents posted on 20 different topics. Various versions of this dataset are used for text classification, text clustering and so one. One of the most popular versions contains 18,821 documents that are evenly classified across all topics.

\paragraph{\textbf{Sogou News}}

The Sogou News dataset~\cite{sun2019fine} is a mixture of the SogouCA and SogouCS news corpora. 
The classification labels of the news are determined by their domain names in the URL. For example, the news with URL http://sports.sohu.com is categorized as a sport class.

\paragraph{\textbf{Reuters news}}
The Reuters-21578 dataset~\cite{reuters} is one of the most widely used data collections for text categorization, and is collected from the Reuters financial newswire service in 1987. 
ApteMod is a multi-class version of Reuters-21578 with 10,788 documents. It has 90 classes, 7,769 training documents and 3,019 test documents. 
Other datasets derived from a subset of the Reuters dataset include R8, R52, RCV1, and RCV1-v2.

Other datasets developed for news categorization includes: Bing news~\cite{wang2014concept}, BBC~\cite{greene06icml}, Google news~\cite{das2007google}.

\subsection{\textbf{Topic Classification Datasets}}

\paragraph{\textbf{DBpedia}}

The DBpedia dataset~\cite{lehmann2015dbpedia} is a large-scale, multilingual knowledge base that has been created from the most commonly used infoboxes within Wikipedia. DBpedia is published every month and some classes and properties are added or removed in each release. The most popular version of DBpedia contains 560,000 training samples and 70,000 test samples, each with a 14-class label.

\paragraph{\textbf{Ohsumed}}

The Ohsumed collection~\cite{ohsumed} is a subset of the MEDLINE database. Ohsumed contains 7,400 documents. Each document is a medical abstract that is labeled by one or more classes selected from 23 cardiovascular diseases categories. 

\paragraph{\textbf{EUR-Lex}}

The EUR-Lex dataset~\cite{mencia2008efficient} includes different types of documents, which are indexed according to several orthogonal categorization schemes to allow for multiple search facilities. 
The most popular version of this dataset is based on different aspects of European Union law and has 19,314 documents and 3,956 categories.

\paragraph{\textbf{WOS}}

The Web Of Science (WOS) dataset~\cite{kowsari2017hdltex} is a collection of data and meta-data of published papers available from the Web of Science, which is the world's most trusted publisher-independent global citation database. WOS has been released in three versions: WOS-46985, WOS-11967 and WOS-5736. WOS-46985 is the full dataset. 
WOS-11967 and WOS-5736 are two subsets of WOS-46985.

\paragraph{\textbf{PubMed}}

PubMed~\cite{lu2011pubmed} is a search engine developed by the National Library of Medicine for medical and biological scientific papers, which contains a document collection. Each document has been labeled with the classes of the MeSH set which is a label set used in PubMed. 
Each sentence in an abstract is labeled with its role in the abstract using one of the following classes: background, objective, method, result, or conclusion.


Other datasets for topic classification includes PubMed 200k RCT~\cite{dernoncourt2017pubmed}, Irony (which is composed of annotated comments from the social news website reddit, Twitter dataset for topic classification of tweets, arXiv collection)~\cite{wallace2014humans}, to name a few.

\subsection{\textbf{QA Datasets}}

\paragraph{\textbf{SQuAD}}

Stanford Question Answering Dataset (SQuAD)~\cite{rajpurkar2016squad} is a collection of question-answer pairs derived from  Wikipedia articles. In SQuAD, the correct answers of questions can be any sequence of tokens in the given text. 
Because the questions and answers are produced by humans through crowdsourcing, it is more diverse than some other question-answering datasets. 
SQuAD 1.1 contains 107,785 question-answer pairs on 536 articles. 
SQuAD2.0, the latest version, combines the 100,000 questions in SQuAD1.1 with over 50,000 un-answerable questions written adversarially by crowdworkers in forms that are similar to the answerable ones~\cite{rajpurkar2018know}.

\paragraph{\textbf{MS MARCO}} 
This dataset is released by Microsoft~\cite{nguyen2016ms}. Unlike SQuAD where all questions are produced by edits; In MS MARCO, all questions are sampled from user queries and passages from real web documents using the Bing search engine. Some of the answers in MS MARCO are generative. So, the dataset can be used to develop generative QA systems. 

\paragraph{\textbf{TREC-QA}} 

TREC-QA~\cite{trecQA} is one of the most popular and studied datasets for QA research. 
This dataset has two versions, known as TREC-6 and TREC-50.
TREC-6 consists of questions in 6 categories while TREC-50 in fifty classes. 
For both versions, the training and test datasets contain 5,452 and 500 questions, respectively.

\paragraph{\textbf{WikiQA}}

The WikiQA dataset~\cite{yang2015wikiqa} consists of a set of question-answer pairs, collected and annotated for open-domain QA research. The dataset also includes questions for which there is no correct answer, allowing researchers to evaluate answer triggering models.

\paragraph{\textbf{Quora}}

The Quora dataset~\cite{quora} is developed for paraphrase identification (to detect duplicate questions). 
For this purpose, the authors present a subset of Quora data that consists of over 400,000 question pairs. 
A binary value is assigned to each question pair indicating whether the two questions are the same or not.

Other datasets for QA includes Situations With Adversarial Generations (SWAG)~\cite{zellers2018swag}, WikiQA~\cite{yang2015wikiqa}, SelQA~\cite{jurczyk2016selqa}.

\subsection{\textbf{NLI Datasets}}

\paragraph{\textbf{SNLI}} 

The Stanford Natural Language Inference (SNLI) dataset~\cite{bowman2015large} is widely used for NLI.
This dataset consists of 550,152, 10,000 and 10,000 sentence pairs for training, development and test, respectively.
Each pair is annotated with one of the three labels: neutral, entailment, contradiction. 

\paragraph{\textbf{Multi-NLI}} 

The Multi-Genre Natural Language Inference (MNLI) dataset~\cite{williams2017broad} is a collection of 433k sentence pairs annotated with textual entailment labels. 
The corpus is an extension of SNLI, covers a wider range of genres of spoken and written text, and supports a distinctive cross-genre generalization evaluation.

\paragraph{\textbf{SICK}} 

The Sentences Involving Compositional Knowledge (SICK) dataset~\cite{marelli2014semeval} consists of about 10,000 English sentence pairs which are annotated with three labels: entailment, contradiction, and neutral.

\paragraph{\textbf{MSRP}} 

The Microsoft Research Paraphrase (MSRP) dataset~\cite{dolan2004unsupervised} is commonly used for the text similarity task. MSRP consists of 4,076 samples for training and 1,725 samples for testing. Each sample is a sentence pair, annotated with a binary label indicating whether the two sentences are paraphrases or not. 


Other NLI datasets includes Semantic Textual Similarity (STS)~\cite{cer2017semeval}, RTE~\cite{Dagan2006}, SciTail~\cite{Khot2018}, to name a few.

\section{\textbf{Experimental Performance Analysis}}
\label{sec:performance}

In this section, we first describe a set of metrics commonly used for evaluating TC models' performance, and then present a quantitative analysis of the performance of a set of DL-based TC models on popular benchmarks.

\subsection{\textbf{Popular Metrics for Text Classification}}


\paragraph{\textbf{Accuracy and Error Rate}} 
These are primary metrics to evaluate the quality of a classification model. 
Let TP, FP, TN, FN denote true positive, false
positive, true negative, and false negative, respectively.
The classification Accuracy and Error Rate are defined in Eq.~\ref{accuracy_error_r}
\begin{equation}
\begin{aligned}
\textbf{Accuracy}= \frac{(\text{TP} + \text{TN})}{N},
\ \ \ \
\textbf{Error rate}= \frac{(\text{FP} + \text{FN})}{N},
\label{accuracy_error_r}
\end{aligned}
\end{equation}
where $N$ is the total number of samples. Obviously, we have \textbf{Error Rate} = \textbf{1 - Accuracy}.

\paragraph{\textbf{Precision\,/\,Recall\,/\,F1 score}} 
These are also primary metrics, and are more often used than accuracy or error rate for imbalanced test sets, e.g., the majority of the test samples have one class label.  
Precision and recall for binary classification are defined as
Eq.~\ref{prec_rec}.
The F1 score is the harmonic mean of the precision and recall, as in Eq.~\ref{prec_rec}. An F1 score reaches its best value at 1 (perfect precision and recall) and worst at 0.

\begin{equation}
\begin{aligned}
\textbf{Precision}&= \frac{\text{TP}}{\text{TP}+\text{FP}}, \ \ \ \
\textbf{Recall}&= \frac{\text{TP}}{\text{TP}+\text{FN}}, \ \ \ \textbf{F1-score}= \frac{2 \ \text{Prec} \ \text{Rec}}{ \text{Prec}+\text{Rec} }
\end{aligned}
\label{prec_rec}
\end{equation}

For multi-class classification problems, we can always compute precision and recall for each class label and analyze the individual performance on class labels or average the values to get the overall precision and recall.

\paragraph{\textbf{Exact Match (EM)}}

The exact match metric is a popular metric for question-answering systems, which measures the percentage of predictions that match any one of the ground truth answers exactly. 
EM is one of the main metrics used for SQuAD.

\paragraph{\textbf{Mean Reciprocal Rank (MRR)}}

MRR is often used to evaluate the performance of ranking algorithms in NLP tasks such as query-document ranking and QA. 
MRR is defined in Eq.~\ref{mrr}, where $Q$ is a set of all possible answers, and $rank_{i}$ is the ranking position of the ground-truth answer.

\begin{equation}
\textbf{MRR}= \frac{1}{|Q|} \sum_{i=1}^{Q} \frac{1}{rank_{i}}.
\label{mrr}
\end{equation}

Other widely used metrics include Mean Average Precision (MAP), Area Under Curve (AUC), False Discovery Rate, False Omission Rate, to name a few.
 

\subsection{\textbf{Quantitative Results}}

We tabulate the performance of several of the previously discussed algorithms on popular TC benchmarks.
In each table, in addition to the results of a set of representative DL models, we also present results using non-deep-learning models which are either previous state of the art or widely used as baselines before the DL era. We can see that across all these tasks, the use of DL models leads to significant improvements.

Table~\ref{table_sentiment} summarizes the results of the models described in Section \ref{sec:Deep_text} on several sentiment analysis datasets, including Yelp, IMDB, SST, and Amazon. 
We can see that significant improvement in accuracy has been obtained since the introduction of the first DL-based sentiment analysis model, e.g., with around 78\% relative reduction in classification error (on SST-2).

Table~\ref{table_news_topic} reports the performance on three news categorization datasets (i.e., AG News, 20-NEWS, Sogou News) and two topic classification datasets (i.e., DBpedia and Ohsummed). 
A similar trend to that in sentiment analysis is observed. 

Tables~\ref{table_squad} and \ref{table_wikiQA} present the performance of some DL models on SQuAD, and WikiQA, respectively.
It is worth noting that on both datasets the significant performance lift is attributed to the use of BERT. 

Table~\ref{table_NLI} presents the results on two NLI datasets (i.e., SNLI and MNLI). We observe a steady performance improvement on both datasets over the last 5 years.

\begin{table}[h]
\centering
\caption{Accuracy of deep learning based text classification models on sentiment analysis datasets (in terms of classification accuracy), evaluated on the IMDB, SST, Yelp, and Amazon datasets. Italic indicates the non-deep-learning models.}
\label{table_sentiment}
\begin{tabular}{llllllr}
\toprule
Method & IMDB & SST-2 & Amazon-2  &  Amazon-5 & Yelp-2 & Yelp-5\\
\midrule
\textit{Naive Bayes}  \cite{socher2013recursive}  & -  & 81.80  & - & - & - & -  \\ \hline
\textit{LDA}  \cite{maas2011learning}  & 67.40  & -  & - & - & - & -  \\ \hline
\textit{BoW+SVM}  \cite{wang2012baselines}  & 87.80  & -  & - & - & - & -  \\ \hline
\textit{tf.$\Delta$ idf}  \cite{martineau2009delta}  & 88.10  & -  & - & - & - & -  \\ \hline
Char-level CNN  \cite{zhang2015character}  & -  & -  & 94.49 & 59.46 & 95.12 & 62.05  \\ \hline  
Deep Pyramid CNN  \cite{Johnson2017}  & -  & 84.46  & 96.68 & 65.82 &  97.36 & 69.40  \\ \hline  
ULMFiT \cite{howard2018universal}  & 95.40  &  - & - & - & 97.84 & 70.02  \\  \hline 
BLSTM-2DCNN \cite{zhou2016text}  & -  &  89.50 & - & - &- & -\\  \hline 
Neural Semantic Encoder \cite{munkhdalai2017neural}  & -  &  89.70 & - & - &-& -\\ \hline 
BCN+Char+CoVe  \cite{mccann2017learned}  & 91.80  &  90.30 & - & - & - & - \\ \hline 
GLUE ELMo baseline \cite{wang2018glue} &  - &  90.40 & - & -&  -& -\\ \hline 
BERT ELMo baseline \cite{devlin2018bert} & -  &  90.40 & - & -&-  & -\\ \hline 
CCCapsNet \cite{ren2018compositional}  &- & - &  94.96 &  60.95 & 96.48 &65.85 \\ \hline 
Virtual adversarial training \cite{miyato2016adversarial}  & 94.10  & - & - & - &- &  -\\ \hline 
Block-sparse LSTM  \cite{gray2017gpu}  & 94.99  & 93.20 & - & - & 96.73 &  \\ \hline 
BERT-base  \cite{devlin2018bert, sun2019fine} & 95.63  & 93.50 & 96.04 & 61.60 & 98.08 & 70.58 \\ \hline 
BERT-large  \cite{devlin2018bert, sun2019fine} & 95.79  & 94.9 & 96.07 & 62.20 & 98.19 & 71.38 \\ \hline 
ALBERT  \cite{lan2019albert} & - & 95.20 & - &  -& - & - \\ \hline 
Multi-Task DNN \cite{liu2019multi} & 83.20 & 95.60 & - & -& - & - \\ \hline 
Snorkel MeTaL  \cite{ratner2019training} & - & 96.20 & - & - &- & - \\ \hline 
BERT Finetune + UDA \cite{xie2019unsupervised} & 95.80 &  & 96.50 & 62.88 & 97.95 & 62.92 \\ \hline 
RoBERTa (+additional data) \cite{liu2019roberta} & - & 96.40  & - & - & - & - \\ \hline 
XLNet-Large (ensemble) \cite{yang2019xlnet} & 96.21  & 96.80 & 97.60 & 67.74 & 98.45 & 72.20 \\ 
\bottomrule
\end{tabular}
\end{table}

\begin{table}[h]
\centering
\caption{Accuracy of classification models on news categorization, and topic classification tasks. Italic indicates the non-deep-learning models.}
\label{table_news_topic}
\begin{tabular}{lccc|cc}
\hline
&  \multicolumn{3}{c}{\textbf{News Categorization}} & \multicolumn{2}{c}{\textbf{Topic Classification}}\\
\cline{1-6}
Method  & AG News  &   20NEWS  & Sogou News &  DBpedia  &  Ohsumed  \\ \hline 
\textit{Hierarchical  Log-bilinear  Model} \cite{kusner2015word}  & - & - & - & - & 52 \\  \hline
Text GCN \cite{yao2019graph}  & 67.61 & 86.34 & - & - & 68.36 \\  \hline
Simplfied GCN \cite{wu2019simplifying}  & -  &  88.50  & - & -  &  68.50    \\ \hline
Char-level CNN  \cite{zhang2015character} & 90.49  & - & 95.12 & 98.45  & -  \\ \hline
CCCapsNet \cite{ren2018compositional}  & 92.39  & - & 97.25 &  98.72 & -   \\ \hline
LEAM \cite{wang2018joint} & 92.45  & 81.91 & - & 99.02  & 58.58\\ \hline
fastText \cite{joulin2016fasttext}  & 92.50  & - & 96.80 & 98.60  & 55.70 \\ \hline
CapsuleNet B \cite{zhao2018investigating}  & 92.60  & - & - & - & - \\ \hline
Deep Pyramid CNN \cite{Johnson2017}  & 93.13  & - & 98.16 & 99.12  & - \\ \hline
ULMFiT \cite{howard2018universal}  & 94.99  & - & - & 99.20  & - \\ \hline
L MIXED \cite{sachan2019revisiting}  &  95.05  & - & - &  99.30  & - \\ \hline
BERT-large  \cite{xie2019unsupervised}  & - & - & - &  99.32  &  - \\ \hline
XLNet \cite{yang2019xlnet}  & 95.51  & - & - & 99.38  & -\\ 
\bottomrule
\end{tabular}
\end{table}


\begin{table}[h]
\centering
\caption{Performance of classification models on SQuAD question answering datasets. Here, the F1 score  measures the average overlap between the prediction and ground truth answer. Italic denotes the non-deep-learning models.}
\label{table_squad}
\begin{tabular}{lcccc}
\toprule
& \multicolumn{2}{c}{SQuAD1.1} & \multicolumn{2}{c}{SQuAD2.0}\\
\midrule
Method & EM & F1-score & EM & F1-score \\ \midrule
\textit{Sliding Window+Dist.}~\cite{richardson2013mctest} & 13.00	& 20.00 & - & -  \\ 
\textit{Hand-crafted Features+Logistic Regression}~\cite{rajpurkar2016squad} \ \ \  & 40.40	& 51.00 & - & -  \\
BiDAF + Self Attention + ELMo~\cite{peters2018deep} & 78.58	& 85.83 & 63.37 & 66.25  \\ 
SAN (single model)~\cite{liu2017stochastic} & 76.82 &	84.39 & 68.65 & 71.43  \\ 
FusionNet++ (ensemble)~\cite{huang2017fusionnet} & 78.97 &	86.01 & 70.30 &	72.48 \\ 
SAN (ensemble)~\cite{liu2017stochastic} & 79.60 & 86.49 & 71.31 &	73.70 \\ 
BERT (single model)~\cite{devlin2018bert} & 85.08 &	91.83	 & 80.00 &	83.06 \\
BERT-large (ensemble)~\cite{devlin2018bert} & 87.43 & 	93.16 & 80.45 &	83.51 \\
BERT + Multiple-CNN~\cite{liu2017stochastic} & - & - & 84.20 &	86.76\\
XL-Net~\cite{yang2019xlnet} & 89.90 & 95.08 & 84.64 & 88.00 \\
SpanBERT~\cite{joshi2019spanbert} & 88.83 &	94.63 & 71.31 &	73.70 \\
RoBERTa~\cite{liu2019roberta} & - & - & 86.82 & 89.79 \\
ALBERT (single model)~\cite{lan2019albert} & - & - & 88.10	& 90.90 \\
ALBERT (ensemble)~\cite{lan2019albert} & - & - & 89.73 & 92.21\\
Retro-Reader on ALBERT & - & - & 90.11 & 92.58  \\
ELECTRA+ALBERT+EntitySpanFocus  & - & - & 90.42 & 92.79  \\
\bottomrule
\end{tabular}
\end{table}

\begin{table}[h]
\centering
\caption{Performance of classification models on the WikiQA datasets.}
\label{table_wikiQA}
\begin{tabular}{lcc}
\toprule
Method & MAP & MRR \\ \midrule
Paragraph vector~\cite{le2014distributed} & 0.511 & 0.516 \\ 
Neural Variational Inference~\cite{miao2016neural} &  0.655 & 0.674 \\ 
Attentive pooling networks~\cite{santos2016attentive} &  0.688 & 0.695 \\ 
HyperQA~\cite{tay2018hyperbolic} & 0.712	& 0.727 \\
BERT (single model)~\cite{devlin2018bert} & 0.813 & 0.828 \\
TANDA-RoBERTa~\cite{garg2019tanda} & 0.920	& 0.933 \\
\bottomrule
\end{tabular}
\end{table}

\begin{table}[h]
\centering
\caption{Performance of classification models on natural language inference datasets. For Multi-NLI, Matched and Mismatched refer to the matched and mismatched test accuracies, respectively. Italic denotes the non-deep-learning models.}
\label{table_NLI}
\begin{tabular}{lccc}
\toprule
& SNLI & \multicolumn{2}{c}{MultiNLI}\\
\midrule
Method & Accuracy & Matched & Mismatched \\
\midrule
\textit{Unigrams Features} \cite{bowman2015large}  & 71.6  &- & - \\ 
\textit{Lexicalized} \cite{bowman2015large}  & 78.2  &- & - \\ 
LSTM encoders (100D) \cite{bowman2015large}  & 77.6  &- & - \\ 
Tree Based CNN \cite{mou2015natural}  & 82.1  & - & - \\ 
biLSTM Encoder \cite{williams2017broad} & 81.5 & 67.5 & 67.1\\
Neural Semantic Encoders (300D) \cite{munkhdalai2017neural}  &  84.6  & - & - \\
RNN Based Sentence Encoder  \cite{chen2017recurrent}  & 85.5  & 73.2 & 73.6  \\
DiSAN (300D)  \cite{shen2018disan}  & 85.6  & - & - \\
Decomposable Attention Model \cite{parikh2016decomposable}  & 86.3  & - & - \\
Reinforced Self-Attention (300D)  \cite{shen2018reinforced}  & 86.3  & - & - \\
Generalized Pooling (600D)  \cite{chen2018enhancing} & 86.6  &  73.8 & 74.0  \\
Bilateral multi-perspective matching  \cite{wang2017bilateral}  & 87.5  & - & - \\
Multiway Attention Network \cite{tan2018multiway} & 88.3  &  78.5 & 77.7 \\
ESIM + ELMo  \cite{peters2018deep} & 88.7  &  72.9 &	73.4  \\
DMAN with Reinforcement Learning  \cite{pan2019discourse} & 88.8  &  88.8 & 78.9  \\
BiLSTM + ELMo + Attn \cite{wang2018glue} & - & 74.1 & 74.5  \\ 
Fine-Tuned LM-Pretrained Transformer  \cite{radford2018improving} & 89.9  & 82.1 & 81.4  \\
Multi-Task DNN \cite{liu2019multi} & 91.6  &  86.7 & 86.0  \\
SemBERT  \cite{zhang2019semantics} & 91.9  & 84.4 & 84.0   \\
RoBERTa \cite{liu2019roberta} &  92.6 & 90.8 & 90.2 \\
XLNet \cite{yang2019xlnet} &- &90.2 & 89.8\\
\hline
\end{tabular}
\end{table}

\section{\textbf{Challenges and Opportunities}}
\label{sec:challenges}
TC has seen a great progress over the last few years, with the help of DL models. 
Several novel ideas have been proposed (such as neural embedding, attention mechanism, self attention, Transformer, BERT, and XLNet), which lead to the fast progress over the past decade. 
Despite the progress, there are still challenges to be addressed.
This section presents some of these challenges, and discusses research directions that could help advance the field.

\paragraph{\textbf{New Datasets for More Challenging Tasks}}
Although a number of large-scale datasets have been collected for common TC tasks in recent years, there remains a need for new datasets for more challenging TC tasks such as QA with multi-step reasoning, text classification for multi-lingual documents, and TC for extremely long documents.

\paragraph{\textbf{Modeling Commonsense Knowledge}}
Incorporating commonsense knowledge into DL models has a potential to significantly improve model performance, pretty much in the same way that humans leverage commonsense knowledge to perform different tasks.
For example, a QA system equipped with a commonsense knowledge base could answer questions about the real world.
Commonsense knowledge also helps to solve problems in the case of incomplete information. Using widely held beliefs about everyday objects or concepts, AI systems can reason based on ``default'' assumptions about the unknowns in a similar way people do.
Although this idea has been investigated for sentiment classification~\cite{cambria2016senticnet}, much more research is required to explore to effectively model and use commonsense knowledge in DL models. 

\paragraph{\textbf{Interpretable DL Models}}
While DL models have achieved promising performance on challenging benchmarks, most of these models are not interpretable. 
For example, why does a model outperform another model on one dataset, but underperform on other datasets?  
What exactly have DL models learned? 
What is a minimal neural network architecture that can achieve a certain accuracy on a given dataset? 
Although the attention and self-attention mechanisms provide some insight toward answering these questions, a detailed study of the underlying behavior and dynamics of these models is still lacking. 
A better understanding of the theoretical aspects of these models can help develop better models curated toward various text analysis scenarios.

\paragraph{\textbf{Memory Efficient Models}}
Most modern neural language models require a significant amount of memory for training and inference. 
These models have to be compressed in order to meet the computation and storage constraints of edge applications.
This can be done either by building student models using knowledge distillation, 
or by using model compression techniques.
Developing a task-agnostic model compression method is an active research topic~\cite{wang2020minilm}.   

\paragraph{\textbf{Few-Shot and Zero-Shot Learning}}
Most DL models are supervised models that require large amounts of domain labels. 
In practice, it is expensive to collect such labels for each new domain. 
Fine-tuning a PLM (e.g., BERT and OpenGPT) to a specific task requires much fewer domain labels than training a model from scratch, thus opening opportunities of developing new zero-shot or few-shot learning methods based on PLMs.



\section{\textbf{Conclusion}}
\label{sec:conclusions}
In this paper, we survey more than 150 DL models, which are developed in the past six years and have significantly improved state of the art on various TC tasks.
We also provide an overview of more than 40 popular TC datasets, and present a quantitative analysis of the performance of these models on several public benchmarks.
Finally, we discuss some of the open challenges and future research directions.


\begin{acks}
The authors would like to thank Richard Socher, Kristina Toutanova, Brooke Cowan, and all the anonymous reviewers for reviewing this work and providing very insightful comments.
\end{acks}

%

\bibliographystyle{IEEEtran}


%

\newpage
\appendix

\section{\textbf{Deep Neural Network Overview}}
\label{sec:DNNs}
This appendix introduces some of the commonly used deep learning models for NLP, including MLPs, CNNs, RNNs, LSTMs, encoder-decoders, and Transformers.
Interested readers are referred to~\cite{goodfellow2016deep} for a comprehensive discussion.

\subsection{\textbf{Neural Language Models and Word Embedding}}

Language modeling adopts data-driven approaches to capture salient statistical properties of text sequences in natural language, which can later be used to predict future words in a sequence, or to perform slot-filling in related tasks. 
N-gram models are the simplest statistical language models, which capture the relation between successive tokens. 
However, these models cannot capture long-distance dependence of tokens which often encodes semantic relations~\cite{JurafskyNLPBook}. 
Therefore, there have been a lot of efforts of developing richer language models, among which one of the most successful is the neural language model~\cite{bengio2003neural}.

Neural language models learn to represent textual-tokens (such as words) as dense vectors, referred as to word embeddings, in a self-supervised fashion.
These learned representations can then be used for various NLP applications. 
One popular neural language model is word2vec~\cite{mikolov2013efficient}, which learns to map the words that come in similar contexts to similar vector representations. 
The learned word2vec representations also allow for some simple algebraic operations on word embeddings in vector space, 
as shown in Eq.~\ref{eq:analogy_test}.
\begin{equation}\label{eq:analogy_test}
  ``king'' - ``man'' + ``woman'' = ``queen''
\end{equation}

Despite its popularity and semantic richness, word2vec suffers from some problems such as out of vocabulary (OOV) extension, inability to capture word morphology and word context. 
There have been many works trying to improve word2vec model, and depending on the textual units they deal with and whether being context dependent or not, they can be grouped into the following categories:
\begin{itemize}
  \item Word-Level Embedding
  \item Subword Embedding
  \item Contextual Embedding
\end{itemize}

\begin{figure}[h]
  \centering
  \includegraphics[width=8 cm]{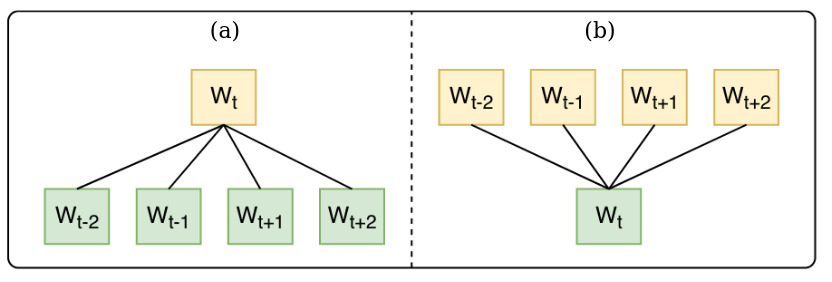}
  \caption{Two word2vec models~\cite{mikolov2013efficient}  (a) CBOW (b) Skip-Gram}
  \label{fig:skip_cbow}
\end{figure}

\paragraph{\textbf{Word-Level Embedding}}
Two main categories of word-level embedding models are prediction-based and count-based models. 
The models in the former category are trained to recover the missing tokens in a token sequence. 
Word2vec is an early example of this category, which proposed two architectures for word embedding, Continuous Bag of Words (CBOW) and Skip-Gram ~\cite{mikolov2013efficient,mikolov2013distributed},
as shown in Fig.~\ref{fig:skip_cbow}.
A Skip-Gram model predicts each context word from the central word, while a CBOW model predicts the central word based on its context words. 
The training objectives of these model are to maximize the prediction probability of the correct words. 
For example, the training objectives of CBOW and Skip-Gram are shown in Eq.~\ref{eq:cbowloss} and Eq.~\ref{eq:skipgramloss}, respectively.
\begin{equation}
  \label{eq:cbowloss}
  \mathcal{L}_{CBOW} = -\frac{1}{|\mathcal{C}|-C}\sum_{k=C+1}^{|\mathcal{C}|-C}\log P(w_k|w_{k-C},\dots, w_{k-1}, w_{k+1}, \dots, w_{k+c})
\end{equation}

\begin{equation}
  \label{eq:skipgramloss}
  \mathcal{L}_{Skip-Gram} =
  - [ \log \sigma({v'_{w}}^\top v_{w_I}) + \sum_{\substack{i=1 \\ \tilde{w}_i \sim Q}}^N \log \sigma(-{v'_{\tilde{w}_i}}^\top v_{w_I})]
\end{equation}

GloVe~\cite{pennington2014glove} is one of the most widely used count-based embedding models. It performs matrix factorization on the co-occurrence matrix of words to learn the embeddings.

\paragraph{\textbf{Subword and Character Embedding}}
Word-level embedding models suffer from problems such as OOV.
One remedy is to segment words into subwords or characters for embeddings.
Character-based embedding models not only can handle the OOV words~\cite{zhang2015character,kim2016character}, 
but also can reduce the embedding model size.
Subword methods find the most frequent character segments (subwords), and then learn the embeddings of these segments. 
FastText~\cite{joulin2016fasttext} is a popular subword embedding model, which represents each word as a bag of character n-grams. 
This is similar to the letter tri-grams used in DSSMs.
Other popular subword tokenizers include byte pair encoding ~\cite{sennrich2015neural}, WordPiece~\cite{schuster2012japanese}, SentencePiece~\cite{Kudo2018}, and so on.

\paragraph{\textbf{Contextual Embedding}}
The meaning of a word depends on its context. For example, the word ``play'' in the sentence ``kid is playing'' has a different meaning from when it is in ``this play was written by Mozart''. 
Therefore, word embedding is desirable to be context sensitive. Neither Word2vec nor Glove is context sensitive. They simply map a word into the same vector regardless of its context.
Contextualized word embedding models, on the other hand, can map a word to different embedding vectors depending on its context.
ELMo~\cite{peters2018deep} is the first large-scale context-sensitive embedding model which uses two LSTMs in forward and backward directions to encode word context. 

\subsection{\textbf{Recurrent Neural Networks (RNNs) and Long Short-Term Memory (LSTM)}} 
\begin{figure}[h]
\begin{center}
  \includegraphics[page=1,width=0.5\linewidth]{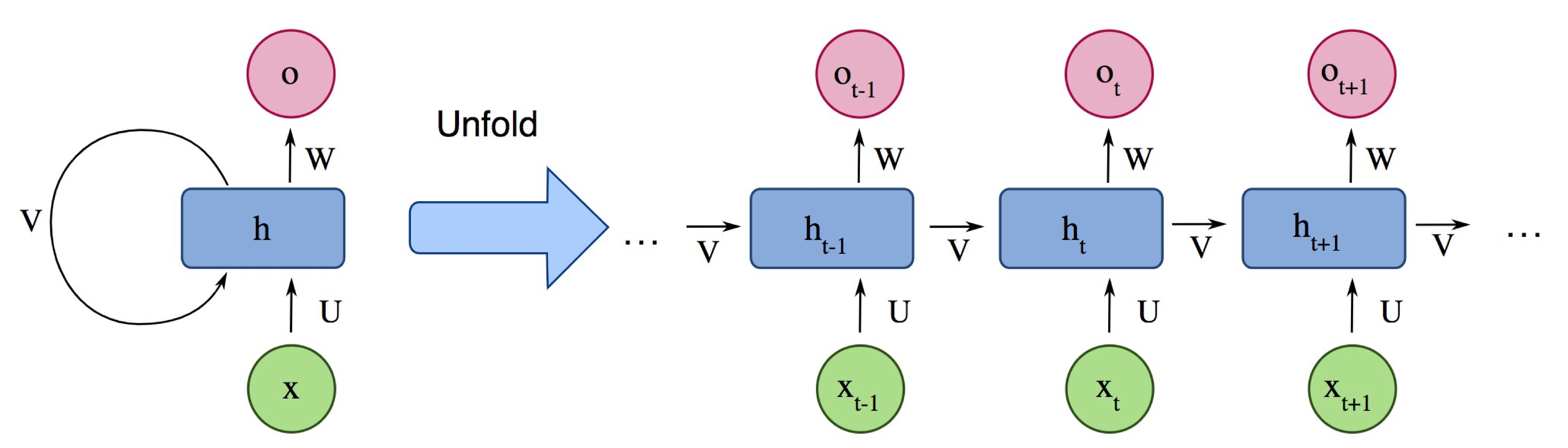} \hspace{1cm}
  \includegraphics[page=14,width=0.22\linewidth]{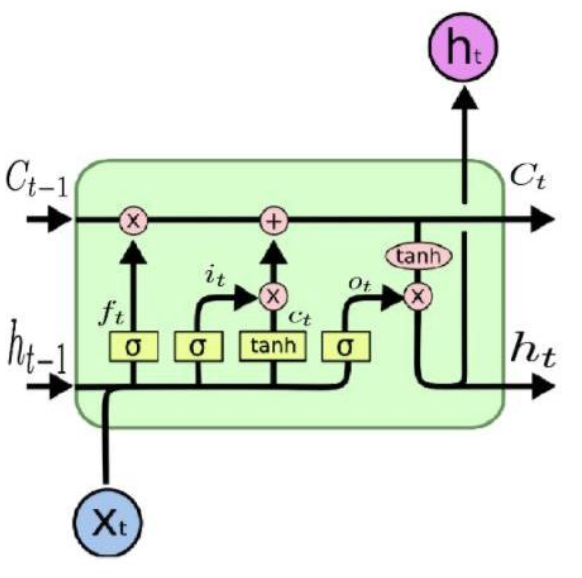}
\end{center}
  \caption{(Left) The architecture of a RNN. (Right) The architecture of a standard LSTM module~\cite{lstm_cell}.}
\label{fig:RNN_arch}
\end{figure}

RNNs~\cite{RNN} are widely used for processing sequential data, such as text, speech, video.
The architecture of a vanilla RNN model is shown in Fig.~\ref{fig:RNN_arch} (left). 
The model gets the input from the current time $X_i$ and the hidden state from the previous step $h_{i-1}$ and generates a hidden state and optionally an output. The hidden state from the last time-stamp (or a weighted average of all hidden states) can be used as the representation of the input sequence for downstream tasks.

RNNs cannot capture long-term dependencies of very long sequences, which appear in many real applications, due to the gradient vanishing and explosion issue. 
LSTM is a variation of RNNs designed to better capture long-term dependencies. 
As shown in Fig.~\ref{fig:RNN_arch} (right) and Eq.~\ref{eq_lstm}, the LSTM layer consists of a memory cell, which remembers values over arbitrary time intervals, and
three gates (input gate, output gate, forget gate) that regulate the flow of information in and out the cell.
The relationship between input, hidden states, and different gates of LSTM is shown in Equation~\ref{eq_lstm}:
\begin{equation}
\begin{aligned}
f_t= \sigma (\textbf{W}^{(f)} x_t+\textbf{U}^{(f)} h_{t-1}+ b^{(f)} ), \hspace{2.73cm} \\
i_t= \sigma (\textbf{W}^{(i)} x_t+\textbf{U}^{(i)} h_{t-1}+ b^{(i)} ), \hspace{2.94cm} \\
o_t= \sigma (\textbf{W}^{(o)} x_t+\textbf{U}^{(o)} h_{t-1}+ b^{(o)} ), \hspace{2.78cm} \\
c_t= f_t \odot c_{t-1}+ i_t \odot \text{tanh} (\textbf{W}^{(c)} x_t+\textbf{U}^{(c)} h_{t-1}+ b^{(c)} ), \hspace{0.11cm} \\
h_t= o_t \odot \text{tanh}(c_t) \hspace{4.88cm}
\end{aligned}
\label{eq_lstm}
\end{equation}
where 
$x_t \in R^k$ is a k-D word embedding input at time-step $t$, 
$\sigma$ is the element-wise sigmoid function, 
$\odot$ is the element-wise product,
\textbf{W}, \textbf{U} and $b$ are model parameters,
$c_t$ is the memory cell,
the forget gate $f_t$ determines whether to reset the memory cell, and
the input gate $i_t$ and output gate $o_t$ control the input and output of the memory cell, respectively.

\subsection{\textbf{Convolutional Neural Networks (CNNs)}} 
\begin{figure}[h]
\begin{center}
  \includegraphics[page=3,width=0.7\linewidth]{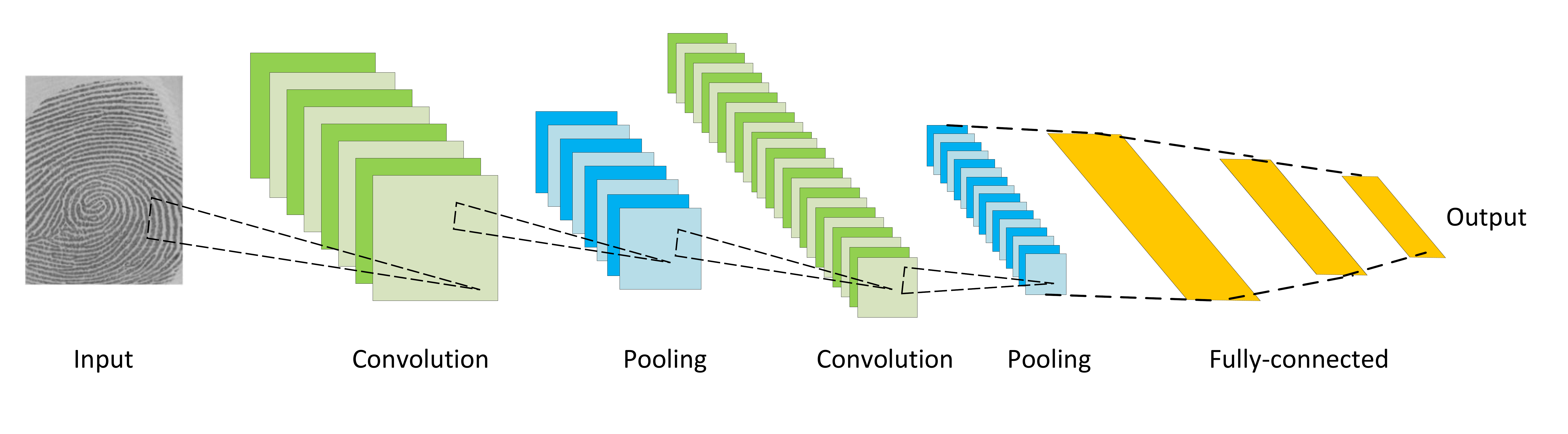}
\end{center}
  \caption{Architecture of a CNN model, courtesy of Yann LeCun~\cite{lecun1998gradient}.}
\label{fig:CNN_arch}
\end{figure}

CNNs are originally developed for computer vision tasks, but later on made their way in various NLP applications.
CNNs were initially proposed by Fukushima in his seminal paper "Neocognitron"~\cite{neocog}, based on the model of the human visual system proposed by Hubel and Wiesel. Yann LeCun and his colleagues popularized CNNs by developing an efficient method of training CNNs based on back-propagation~\cite{lecun1998gradient}.
The architecture of the CNN model developed by LeCun et al. is shown in Fig.~\ref{fig:CNN_arch}.

CNNs consist of three types of layers: (1) the convolutional layers, where a sliding kernel is applied to a region of an image (or a text segment) to extract local features; (2) the nonlinear layers, where a non-linear activation function is applied to (local) feature values; and (3) the pooling layers, where local features are aggregated (via the max-pooling or mean-pooling operation) to form global features. 
One advantage of CNNs is the weight sharing mechanism due to the use of the kernels, which results in a significantly smaller number of parameters than a similar fully-connected neural network, making CNNs much easier to train.
CNNs have been widely used in computer vision, NLP, and speech recognition problems~\cite{alexnet, he2016deep, minaee2020image, Kalchbrenner2014, abdel2014convolutional, minaee2019biometric, gehring2017convolutional}.


\subsection{\textbf{Encoder-Decoder Models}} 
\begin{figure}[h]
\begin{center}
  \includegraphics[page=1,width=0.5\linewidth]{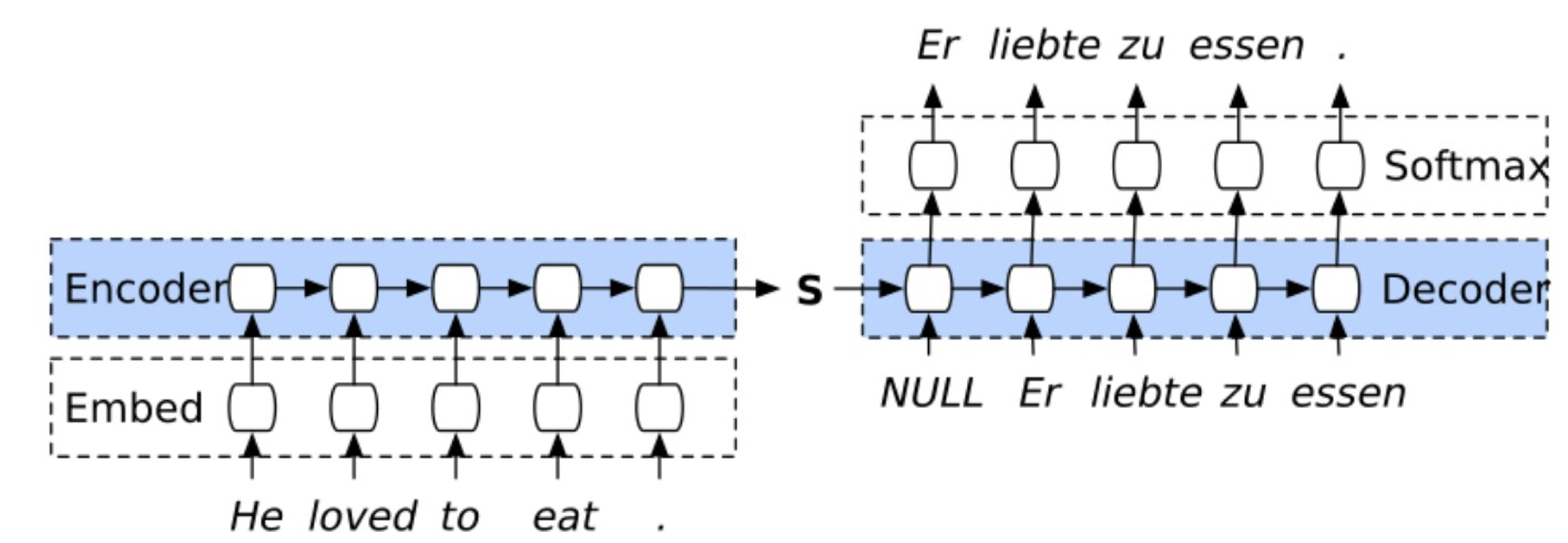}
\end{center}
  \caption{A simple encoder-decoder model for machine translation. The input is a sequence of words in English, and the output is its translated version in German.}
\label{fig:autoencoder}
\end{figure}

Encoder-Decoder models learn to map input to output via a two-stage process: (1) the encoding stage, where an encoder $f(.)$ compresses input $x$ into a latent-space vector representation $z$ as $z=f(x)$; and (2) the decoding stage, where a decoder $g(.)$ reconstructs or predicts output $y$ from $z$ as $y=g(z)$.
The latent representation $z$ is expected to capture the underlying semantics of the input.
These models are widely used in sequence-to-sequence tasks such as machine translation, as illustrated in Fig.~\ref{fig:autoencoder}. 

Autoencoders are special cases of the encoder-decoder models in which the input and output are the same. 
Autoencoders can be trained in an unsupervised fashion by minimizing the reconstruction loss.


\subsection{\textbf{Attention Mechanism}} 

Attention is motivated by how we pay visual attention to different regions of an image or correlate words in one sentence. Attention becomes an increasingly popular concept and useful tool in developing deep learning models for NLP \cite{bahdanau2014neural,luong2015effective}.
In a nutshell, attention in language models can be interpreted as a vector of importance weights. 
In order to predict a word in a sentence, using the attention vector, we estimate how strongly it is correlated with, or ``attends to'', other words and take the sum of their values weighted by the attention vector as the approximation of the target.

\begin{figure}[h]
\begin{center}
  \includegraphics[page=1,width=0.21\linewidth]{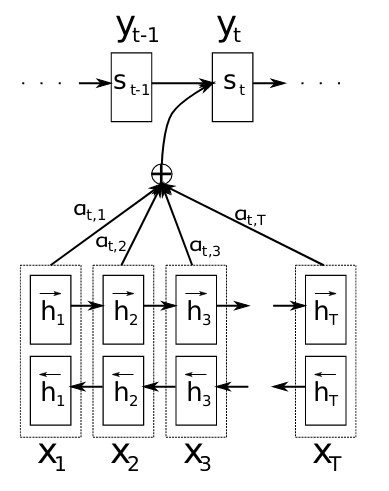} \hspace{2cm}
  \includegraphics[page=1,width=0.27\linewidth]{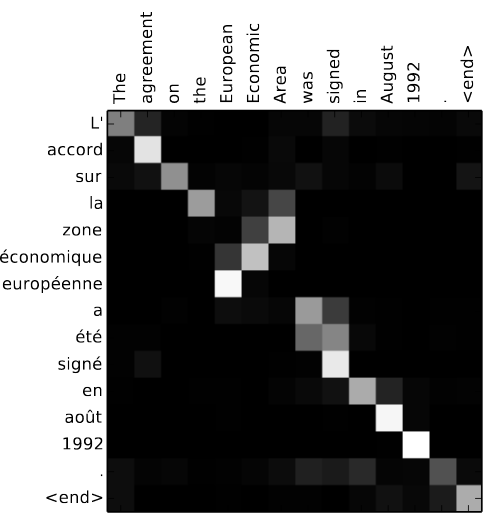}
\end{center}
  \caption{(Left) The proposed attention mechanism in~\cite{bahdanau2014neural}. (Right) An example of attention mechanism in French to English machine translation, which shows the impact of each word in French in translating to English, Brighter cells have more impact.}
\label{fig:attention_bengio}
\end{figure}

Bahdanau et al. ~\cite{bahdanau2014neural} conjectured that the use of a fixed-length state vector in CNNs is the bottleneck in improving the performance of the encoder-decoder model, and proposed to allow the decoder to search for parts in a source sentence that are relevant to predicting the target word, without having to compress the source sentence into the state vector. 
As shown in Fig.~\ref{fig:attention_bengio} (left), a linear combination of hidden vectors of input words $h$, weighted by attention scores $\alpha$, is used to generate the output $y$.
As we can see from Fig.~\ref{fig:attention_bengio} (right), different words in the source sentence are attended with different weights when generating a word in the target sentence.


Self-attention is a special attention mechanism, which allows to learn the correlation among the words in the same sentence~\cite{cheng2016long}. This is very useful in NLP tasks such as machine reading, abstractive summarization, and image captioning.
Transformers, which will be described later, also use self-attention.




\subsection{\textbf{Transformer}}
\label{sec:transformer}

One of the computational bottlenecks suffered by RNNs is the sequential processing of text. 
Although CNNs are less sequential than RNNs, the computational cost to capture meaningful relationships between words in a sentence also grows with increasing length of the sentence, similar to RNNs.
Transformers~\cite{vaswani2017attention} overcome this limitation by 
computing in parallel for every word in a sentence or document an ``attention score'' to model the influence each word has on another.
Due to this feature, Transformers allow for much more parallelization than CNNs and RNNs, and make it possible to efficiently train very big models on large amounts of data on GPU clusters. 

\begin{figure}[h]
\begin{center}
  \includegraphics[page=1,width=0.7\linewidth]{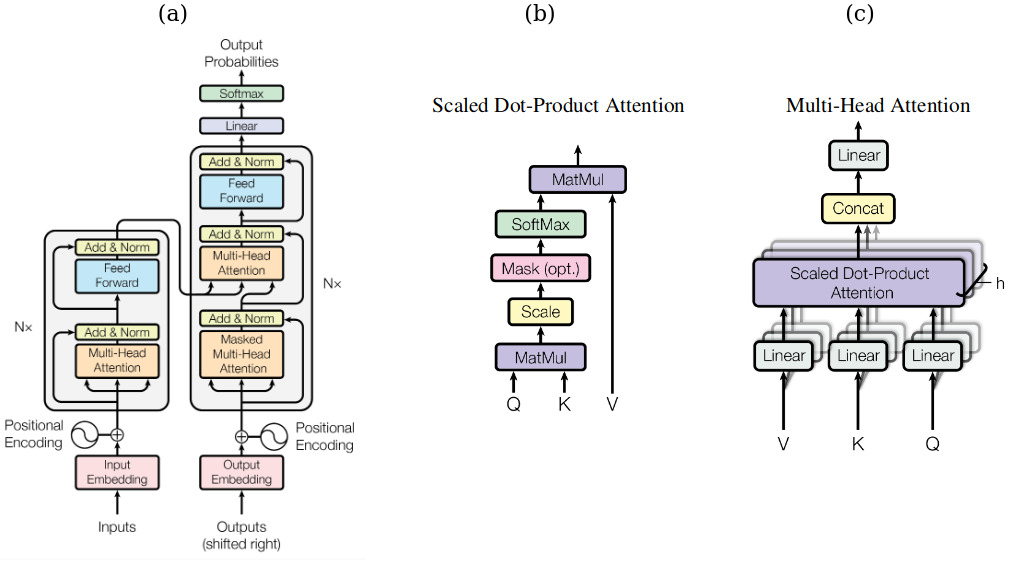}
\end{center}
  \caption{(a) The Transformer model architecture. (b) Scaled Dot-Product Attention. (3) Multi-Head Attention consists of several attention layers running in parallel. \cite{vaswani2017attention}}.
\label{fig:transformer}
\end{figure}

As shown in Fig.~\ref{fig:transformer} (a), the Transformer model consists of stacked layers in both encoder and decoder components.
Each layer has two sub-layers comprising a multi-head attention layer (Fig.~\ref{fig:transformer} (c)) followed by a position-wise feed forward network.
For each set of queries $Q$, keys $K$ and values $V$, the multi-head attention module performs attention $h$ times using the scaled dot-product attention as in Fig.~\ref{fig:transformer} (b), where Mask (option) is the attention mask that is applied to prevent the target word information to be predicted from leaking to the decoder (during training) before prediction.
Experiments show that multi-head attention is more effective than single-head attention. The attention of multiple heads can be interpreted as each head processing a different subspace at a different position. Visualization of the self-attention of multiple heads reveal that each head processes syntax and semantic structures~\cite{vaswani2017attention}.

\end{document}